\documentclass[pdflatex,sn-nature]{sn-jnl}

\usepackage{graphicx}%
\usepackage{multirow}%
\usepackage{amsmath,amssymb,amsfonts}%
\usepackage{amsthm}%
\usepackage{mathrsfs}%
\usepackage{fix-cm}
\usepackage[title]{appendix}%
\usepackage{longtable} 
\usepackage{xcolor}%
\usepackage{textcomp}%
\usepackage{manyfoot}%
\usepackage{booktabs}%
\usepackage{algorithm}%
\usepackage{algorithmicx}%
\usepackage{algpseudocode}%
\usepackage{listings}%
\usepackage{tcolorbox}
\newtheorem*{note}{Note}
\usepackage{caption}
\graphicspath{ {./figures/} }

\theoremstyle{thmstyleone}%
\theoremstyle{thmstyletwo}%
\theoremstyle{thmstylethree}%

\begin{document}

\title[Article Title]{Age Predictors Through the Lens of Generalization, Bias Mitigation, and Interpretability: Reflections on Causal Implications}


\author*[1]{\fnm{Debdas} \sur{Paul}}\email{debdas.paul@leibniz-fli.de}

\author[]{\fnm{Elisa} \sur{Ferrari}}\email{elisa.ferrari.personal@gmail.com}

\author[1,2]{\fnm{Irene} \sur{Gravili}}\email{irene.gravili@leibniz-fli.de}

\author[1,2]{\fnm{Alessandro} \sur{Cellerino}}\email{alessandro.cellerino@sns.it}

\affil[1]{\orgdiv{Leibniz Institute on Aging — Fritz Lipmann Institute (FLI)}, \orgaddress{\street{Beutenbergstr. 11}, \city{Jena}, \postcode{100190}, \country{Germany}}}

\affil[2]{\orgdiv{BIO@SNS}, \orgname{Scuola Normale Superiore}, \orgaddress{\street{Piazza dei Cavalieri,7}, \city{Pisa}, \postcode{10587}, \country{Italy}}}


\abstract{
Chronological age predictors often fail to achieve out-of-distribution (OOD) generalization due to exogenous attributes such as race, gender, or tissue. Learning an invariant representation with respect to those attributes is therefore essential to improve OOD generalization and prevent overly optimistic results. In predictive settings, these attributes motivate bias mitigation; in causal analyses, they appear as confounders; and when protected, their suppression leads to fairness. We coherently explore these concepts with theoretical rigor and discuss the scope of an interpretable neural network model based on adversarial representation learning. Using publicly available mouse transcriptomic datasets, we illustrate the behavior of this model relative to conventional machine learning models. We observe that the outcome of this model is consistent with the predictive results of a published study demonstrating the effects of Elamipretide on mouse skeletal and cardiac muscle. We conclude by discussing the limitations of deriving causal interpretation from such purely predictive models.
}



\maketitle
\newpage
\section*{List of mathematical notations}\label{note_term}
For any scientific dissemination, we believe that it is essential to clarify notations and terminologies in advance. In this way we can avoid incorrect conclusions, misinterpretations, imprecision, and provide clarity for the reader.

\begin{longtable}{ll}
\caption{List of notations used in this work}\label{tab:notations} \\
\toprule
\textbf{Notation} & \textbf{Description} \\
\midrule
\endfirsthead

\multicolumn{2}{c}{\tablename\ \thetable\ -- continued from previous page} \\
\toprule
\textbf{Notation} & \textbf{Description} \\
\midrule
\endhead

\midrule
\multicolumn{2}{l}{continued on next page} \\
\endfoot

\bottomrule
\endlastfoot

$\mathbb{R}$ & Set of real numbers \\
$\mathbb{R}_{\ge 0}$ & Set of nonnegative real numbers \\
$\mathbb{Z}$ & Set of integers \\
$\lceil x \rceil$ & smallest integer greater than or equal to $x$\\
$(\Omega,\mathcal{F},\mathbb{P})$ & Underlying probability space \\
$\mathcal{X}$ & Input (feature) space \\
$\mathcal{Y}$ & Output (label) space \\
$\mathcal{N}(\boldsymbol{\mu},\boldsymbol{\Sigma})$ & Multivariate normal distribution on $\mathbb{R}^d$ \\
$X$ & Input random variable, $X:\Omega\to\mathcal{X}$ \\
$Y$ & Output random variable, $Y:\Omega\to\mathcal{Y}$ \\
$x,y$ & Realizations of random variables $X$ and $Y$ \\
$(X,Y)$ & Joint random variable $(X,Y):\Omega\to\mathcal{X}\times\mathcal{Y}$ \\
$p_{X,Y}$ & Joint probability measure or density on $\mathcal{X}\times\mathcal{Y}$ \\
$p_{Y\mid X}$ & Conditional distribution of $Y$ given $X$ \\
$\mathbb{E}[X]$ & Expectation of $X$ \\
$X \perp\!\!\!\perp Y$ & $X$ and $Y$ are statistically independent \\
$X \not\!\perp\!\!\!\perp Y$ & $X$ and $Y$ are statistically dependent \\
$X \perp\!\!\!\perp Y \mid Z$ & Conditional independence given $Z$ \\
$\mathbb{E}[Y\mid X]$ & Conditional expectation \\
$\mathbb{E}[Y\mid X]=\mathbb{E}[Y]$ & Mean independence of $Y$ from $X$ \\
$\mathcal{P}$ & Data-generating distribution on $\mathcal{X}\times\mathcal{Y}$ \\
$(x_i,y_i)$ & Sample drawn from $\mathcal{D}$ \\
$\{(x_i,y_i)\}_{i=1}^n$ & Dataset of size $n$ \\
$(x_i,y_i)\stackrel{\text{i.i.d.}}{\sim}\mathcal{P}$ & Independent and identically distributed samples \\
$\mathcal{S}$ & Training dataset \\
$f$ & Prediction function $f:\mathcal{X}\to\mathcal{Y}$ \\
$f_\theta$ & Parameterized predictor with $\theta\in\Theta$ \\
$\mathcal{F}$ & Hypothesis class, $\mathcal{F}\subseteq\mathcal{Y}^{\mathcal{X}}$ \\
$\Theta$ & Parameter space \\
$\ell$ & Loss function $\ell:\mathcal{Y}\times\mathcal{Y}\to\mathbb{R}_{\ge 0}$ \\
$R(f)$ & Population risk \\
$\hat{R}(f)$ & Empirical risk on $\mathcal{S}$ \\
$f^*$ & Bayes-optimal predictor \\
$\hat{f}$ & Empirical risk minimizer \\
$\mathrm{do}(X=x)$ & Intervention setting $X$ to $x$~\cite{pearl2009causal} \\
$p_{Y\mid (X=do(x))}$ & Interventional distribution~\cite{pearl2009causal} \\
$\mathbb{S}$ & Set of sample attributes distinct from input features \\
$s \in \mathbb{S}$ &  Attribute $s$ belongs to $\mathbb{S}$\\
$\mathbb{S}_{bio} \subseteq \mathbb{S}$ & Set of Biological attributes (e.g., tissue, strain, sex) \\
$\mathbb{S}_{exp} \subseteq \mathbb{S}$ & Set of Experimental attributes (e.g., platform, protocol, cohort) \\
$\mathbb{S}_{prot} \subseteq \mathbb{S}$ & Set of Protected attributes (e.g., sex, race, ethnicity)
\end{longtable}

\section*{List of Definitions}
\begin{description}

\item[Attribute:]
An \textbf{attribute} refers to contextual information associated with a sample that is distinct from the primary feature vector used for prediction. We denote by $\mathbb{S}$ the set of such attributes and by $s \in \mathbb{S}$ an individual attribute. These attributes may represent biological factors (e.g., tissue or strain), experimental conditions (e.g., sequencing platform or protocol), or protected variables relevant for fairness analysis.
\vspace{10pt}
\item[Sensitive attribute:]
A \textbf{sensitive attribute}, also referred to as a \textbf{protected variable}, is an attribute whose use in predictive models may lead to \textbf{discriminatory outcomes} for particular groups. Sensitive attributes form a subset $\mathbb{S}_{prot} \subseteq \mathbb{S}$. Examples include race, gender, ethnicity, and religion. In fairness-aware learning, models are often designed so that the learned representation is invariant to these attributes.
\vspace{10pt}
\item[Dataset bias:]
\textbf{Dataset bias} refers to systematic dependencies between the learned representation and attributes in $\mathbb{S}$ that arise from heterogeneous biological or experimental environments (e.g., tissue type or sequencing platform). Such dependencies may degrade out-of-distribution generalization. Methods that attempt to remove or reduce these dependencies are commonly referred to as \textbf{bias mitigation} techniques.
\vspace{10pt}
\item[Confounder:]
A \textbf{confounder} is a variable $Z$ that \emph{causally} influences both the independent variable $X$ and the dependent variable $Y$. In such cases the observed statistical relationship between $X$ and $Y$ does not reflect the true causal effect. Formally, in a probabilistic setting
\[
p_{Y \mid X=x} \neq p_{Y \mid do(X=x)}.
\]
Under an appropriate causal model, some attributes in $\mathbb{S}$ may act as confounders when they causally affect both the observed features and the prediction target.
\vspace{10pt}
\item[Bias mitigation:]
\textbf{Bias mitigation} refers to learning strategies that reduce undesirable statistical dependence between the learned representation and attributes in $\mathbb{S}$. The goal is to encourage representations that capture stable predictive signals rather than dataset-specific correlations, thereby improving robustness and out-of-distribution generalization.
\end{description}

\begin{note}
Throughout the manuscript, we refer to machine-learning models that use \emph{chronological age as the target variable} as \textbf{age predictors} or \textbf{chronological age predictors}. This terminology is adopted to avoid ambiguity and to maintain clarity in both mathematical and machine-learning contexts. We do not prefer to use the term "clock" because we believe this should be analogous to features which describe the aging process - the time evolution. But in the community term “clock” is used analogous to a machine learning model, which is ambiguous.  By the word \emph{\textbf{attributes}}, we mean elements in $\mathbb{S}$, unless specified explicitly. 
\end{note}

\tableofcontents
\section{Introduction}\label{sec1}
The problem of chronological age prediction is learning a function $f:\mathcal{X}\rightarrow\mathcal{Y}$, where $\mathcal{X}$ is the input feature space, where features can be genes, proteins, CpGs etc. and $\mathcal{Y}$ is the \emph{chronological age}. This is an example of \emph{supervised learning}, where the learner is presented with the target labels, which is the set $\mathcal{Y}$. The choice of $f$ is governed by two induction principles (rules or assumptions that justify generalizing from finite observations to unseen cases):\emph{Empirical Risk Minimization} (ERM) and \emph{Structural Risk Minimization} (SRM)~\cite{vapnik1991principles}. The term \textbf{Risk} comes from statistical decision theory which measures the cost incurred by the learner in doing prediction. The former one seeks the best fitting function to the \emph{training data}, while the latter one controls the problem of overfitting. The above formulation of the age prediction problem has marked a notable success in the form of famous \emph{epigenetic clock} pioneered by Steve Horvath~\cite{horvath2013dna}. Horvath identified a striking association between DNA methylation patterns and chronological age. However, it is now well understood within the research community that Horvath’s clock does not provide causal information; rather, it is based on statistical correlation. This conclusion is not derived purely from theoretical considerations but is supported by biological knowledge of the underlying processes governing DNA methylation changes during aging. In what follows, we examine why a causal interpretation cannot be established from this association in theory, and under what assumptions such an interpretation might become plausible. Furthermore, Horvath's age predictor has already shown a remarkable predictive performance across different external test cohorts. However, there are instances where the performance is suboptimal, for example on ethnicities other than the training populations (distinct data-generating processes and environments)~\cite{cruz2024methylation}. Despite these nuances, Horvath's age predictor is still considered to be a benchmark. Horvath’s predictor can be viewed as a representative example of a broader class of linear models trained to predict chronological age, irrespective of the specific biological modality from which the features are derived. From a machine learning perspective, such predictors provide a tractable and interpretable setting for analyzing the capabilities and limitations of models that rely on chronological age as the supervisory signal. We therefore use this class of predictors as a conceptual lens to investigate the limits of age prediction and the conditions under which such models succeed or fail. Building on the preceding conceptual analysis, we revisit an adversarial representation learning framework based on deep neural networks~\cite{ferrari2022deep}, providing theoretical grounding for why such formulations arise naturally when learning representations invariant to sample attributes in heterogeneous biological settings. The framework employs an $l_1$-based filtering layer that enables interpretable feature attribution. The adversarial objective suppresses attribute-specific information while predicting chronological age, encouraging representations that are less dependent on contextual attributes. By reducing reliance on spurious correlations induced by such attributes, the resulting representations improve out-of-distribution (OOD) generalization across datasets and experimental conditions (see Figure~\ref{fig:figure3}). We use publicly available mouse transcriptomic datasets (see Materials and Methods in Section~\ref{sec:sec10}) to examine the behavior of the adversarial framework and to analyze its scope and limitations relative to conventional machine learning models. The clinical relevance of any age predictor model is determined by its ability to detect the effect of an intervention (lifestyle or pharmacological) that leaves a molecular signature (epigenetic, transcriptomic, proteomic, etc.), using the same type of data on which the model is trained. To investigate this, we selected a published study~\cite{mitchell2025mitochondria} in which the authors demonstrate the effects of Elamipretide on mouse skeletal and cardiac muscle. When we applied the trained model based on the adversarial framework discussed above to this dataset, we observed two things. First, the model was able to differentiate between the baseline groups (control vs. treatment) in all cases, whereas the conventional models failed in at least one case (see Figure~\ref{fig:figure4}A). Second, the model indicated rejuvenation, which was reported in the study using their in-house age predictor model and is also consistent with other conventional models (see Figure~\ref{fig:figure4}B). This observation suggests the potential value of an ensemble approach in the future. We conclude by discussing the possible scope of causal implications (if any) of such predictive models. We also reignite the discussion on the need for approaches with fully data-driven causal implications in the future, and adversarial representation learning may represent a step in that direction.

In the subsequent sections, we walk through relevant theories to reveal the connections between the concepts of generalization, invariance, bias mitigation, fairness, and interpretability, and how this discussion leads to the adversarial framework presented in this manuscript. These theories are related, but to the best of our knowledge they have not been discussed together in a coherent and systematic manner, particularly in the context of age predictors.

\section{Learning in a heterogeneous environment $\mathcal{E}$}\label{sec:sec2}
Let us consider a prediction problem $f:\mathcal{X}\rightarrow\mathcal{Y}$ under multiple environments $e\in \mathcal{E}$, i.e., $f:\mathcal{X}^{e}\rightarrow\mathcal{Y}^{e}$. Let $\mathcal{F}\supset \mathcal{E}$ denote a (possibly larger) set of environments, and the goal is to generalize to previously unobserved $e\in \mathcal{F}\setminus \mathcal{E}$. In age prediction, such environments may correspond to heterogeneous experimental settings, distinct cohorts, tissues, or population-specific data-generating processes, each of which can induce distribution shift in the observed features.

According to \cite{buhlmann2020invariance}, a common starting point in the invariance literature is to assume:
\begin{itemize}
   \item[(A1)] \label{assump:no_direct_env_effect}
    The environment $e$ does not directly affect the response, i.e., there is no causal arrow $e\rightarrow Y$.
    \item[(A2)] \label{assump:invariant_mechanism}
    The conditional mechanism relating covariates to the response is stable across environments. In practice, this is often more plausible for a subset of features or a learned representation $Z=\Phi(X)$ than for the full covariate vector, i.e., $p_{Y^{e}\mid Z^{e}}$ is invariant across $e$.
\end{itemize}

Under such stability assumptions, a linear predictor can be motivated via a worst-case (robust) $L_2$ risk objective~\cite{buhlmann2020invariance}:
\begin{equation}
\label{eq:worst_case_risk}
\hat{\beta}=\arg\min_{\beta}\; \max_{e \in \mathcal{F}}
\; \mathbb{E}\!\left[ \left( Y^{e} - X^{e}\beta \right)^{2} \right],
\end{equation}
where the expectation is taken under the joint distribution of $(X^{e},Y^{e})$ in environment $e$. Equation~\eqref{eq:worst_case_risk} formalizes \emph{predictive robustness} under distributional heterogeneity. Importantly, links to causal structure arise only under additional assumptions: invariance can, in certain settings, identify predictors that align with causal parents of $Y$, but robustness to unseen environments should not be conflated with interventional causality. From this viewpoint, the cross-tissue training strategy underlying Horvath’s epigenetic clock can be seen as encouraging the use of features whose predictive relationship with age is relatively stable across tissues; this is consistent with (but does not by itself prove) an invariance-based explanation for its empirical robustness. More broadly, causal understanding remains essential in aging research because it concerns how biological systems respond to perturbations and interventions that may alter aging trajectories, which goes beyond predictive stability alone.
\subsection{From invariance to causality}
The following are the two invariance assumption~\cite{buhlmann2020invariance}, which provides a \textbf{sense} of causality to Equation~\ref{eq:worst_case_risk} under linear model assumptions:

\begin{itemize}
\item ($\mathcal{E}$): $\exists$ a subset $S^{*}$ of the covariate indices (including the empty set) such that $$p_{Y^e\mid X^{e}_{S^{*}}}\hspace{2pt} \text{is same}\hspace{2pt} \forall e \in \mathcal{E}$$ 
\item ($\mathcal{F}$): There exists a subset $S^\ast$ and a regression coefficient vector $\beta^\ast$
with support
\[
\operatorname{supp}(\beta^\ast)
= \{\, j \mid \beta^\ast_j \neq 0 \,\}
= S^\ast,
\]
such that, for all $e \in \mathcal{E}$,
\begin{equation}
Y^{e} = X^{e}\beta^\ast + \varepsilon^{e},
\end{equation}
where the error term $\varepsilon^{e}$ is independent of $X^{e}$ and
\[
\varepsilon^{e} \sim F_{\varepsilon},
\]
with $F_{\varepsilon}$ denoting the same distribution for all environments $e$.
\end{itemize}
Now suppose we posit a \emph{structural equation model} (SEM) describing the relationship between the random variables $X$ and $Y$ that satisfies assumptions (A1)–(A2). Under these assumptions, and provided suitable identifiability conditions hold, the direct causal parents of $Y$ satisfy the corresponding invariance property; see Proposition~1 in \cite{buhlmann2020invariance}. In particular, if there exists a subset $S^\ast$ such that the conditional distribution $p_{Y^e \mid X^e_{S^\ast}}$ remains invariant across environments, then $S^\ast$ can be interpreted as the set of causal parents of $Y$ within the assumed SEM. However, applying this interpretation to chronological age prediction requires caution. The invariance-based causal interpretation presupposes a meaningful SEM in which the covariates are direct causes of the response variable. In the case of Horvath’s age predictor, such a structural model is biologically implausible: chronological age represents elapsed time rather than an outcome generated by molecular mechanisms. The biologically plausible causal direction runs from age to methylation changes, not the reverse. Consequently, even if an invariant relationship between methylation features and chronological age is observed across tissues or environments, this invariance should be interpreted as reflecting stable statistical regularities induced by age-driven molecular processes, rather than evidence that methylation features are causal parents of chronological age.

\begin{figure}\label{fig:causal}
\centering
	\includegraphics[scale=0.5]{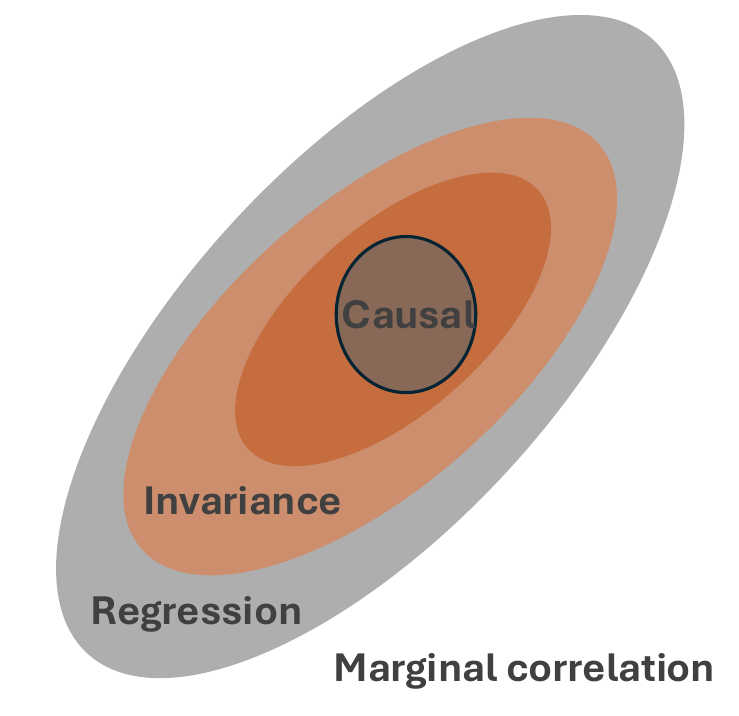}
	\caption{\textbf{Different associations between $X$ and $Y$ as adopted from Figure~12 of \cite{buhlmann2020invariance}}. A marginal correlation is the weakest form association that ignores dependencies among covariates. A stronger form is the regression relevant coefficients which captures partial correlation (non-zero correlation coefficients). Causal variables are a subset of regression variables under faithfulness assumption (A must non-zero coefficient of all parents (direct cause) of $Y$. Invariance set is identifiable even when parents of $Y$ not forming a diluted notion of causality. This set is much more useful in practical applications as stated in \cite{buhlmann2020invariance}. Causal form of association is desirable but often hard to achieve.}
	\label{fig:panel1}
\end{figure}

\begin{tcolorbox}[colback=gray!5,colframe=gray!50]
\textbf{Takeaway}: Invariance across heterogeneous environments can provide a principled route to predictive robustness, particularly under shifts in the marginal distribution of covariates that leave the underlying conditional mechanism stable. However, when chronological age is used as the target variable, such stability should not be conflated with causality. Chronological age is an index of elapsed time rather than an outcome generated by molecular features, and the biologically plausible causal direction typically runs from age to molecular changes rather than the reverse. Consequently, even predictors that generalize robustly across diverse environments should be interpreted as capturing invariant statistical regularities in age-associated biology. Predictive stability under distributional shift does not, by itself, establish causal effects, which would require additional structural assumptions or explicit interventional evidence.
\end{tcolorbox}
In fact, without any causal assumption (or interventional validation), we can not expect any such age predictor will provide a causal answer to questions like "what if and why".  There is currently limited evidence that CpGs used in epigenetic clocks represent direct causal drivers of aging~\cite{ying2024causality}. A good way to summarize this discussion is Figure~\ref{fig:panel1} , which provides a pictorial description of different associations between $X$ and $Y$.
\subsection{Reichenbach's Common Cause Principle (RCCP) - a twist in the tale}
Learning based on correlation implies $X \not\!\perp\!\!\!\perp Y$. Reichenbach’s Common Cause Principle states that such a dependence must arise from an underlying causal structure: either $X$ causes $Y$, $Y$ causes $X$, or there exists a common cause $Z$ influencing both \cite{reichenbach1991direction}. In the context of age prediction, the direction $X \to Y$ (molecular profile causing chronological age) lacks biological plausibility. A more realistic explanation is that chronological age, representing elapsed time, influences molecular states, i.e.,
\[
Y \;\longrightarrow\; X,
\]
through accumulated biological processes. Alternatively, latent biological factors such as cellular turnover, metabolic regulation, or damage accumulation may act as common causes $Z$ that jointly influence both age-associated molecular patterns and phenotypic aging outcomes.  Thus, the correlation exploited by age predictors is more plausibly explained by age-driven or shared biological processes rather than molecular features being causal determinants of age itself. This perspective aligns with our earlier invariance discussion: invariance identifies features that stably track age across environments, but such stability reflects consistent consequences of aging rather than its causes. RCCP becomes interesting when it connects statistical dependence to causality through the possibility of $Z$. In modern causal terminology, such a common cause corresponds to a \emph{confounder}.

\section{Confounding and confounders - correlation is not causation}

Confounding is fundamentally a causal concept in which a variable, called cofounder, causally influences both the predictor and the outcome, thereby inducing a non-causal association~\cite{ye2024spurious} between them. While the term is sometimes used in a purely statistical sense, in this work we adopt the structural (causal) definition unless stated otherwise. An imprecise use of the term risks introducing conceptual ambiguity and undermining the interpretability of the model. For more definitions interested readers are referred to~\cite{vanderweele2013definition}. Presence of confounders (observed/unobserved) is a shortcoming in observational studies~\cite{hernan2020causal} such as ours. But why is this important in our context? At the beginning, we discussed two induction principles namely ERM and SRM following which function learning takes place. One of the main reasons that the predictive model is sensitive to such spurious correlation is due to ERM. ERM fails to distinguish causal vs non-causal signals because it minimizes average risk. ERM may uses a non-causal \textbf{shortcut} path to provide better accuracy for the test set sharing same distribution as that of the training~\cite{ye2024spurious}, but fails for an external dataset where this shortcut is missing. 

Such failures often arise under distribution shifts, particularly when the conditional mechanism changes (concept shift, $p^{e}_{Y \mid X} \neq p^{e'}_{Y \mid X}$). While confounding is one possible source of spurious associations, other mechanisms such as selection bias or reverse causation may also induce non-causal dependencies. The discrepancy between high in-distribution performance and poor robustness under distribution shift therefore motivates learning principles that go beyond ERM and explicitly seek stable relationships across environments, as emphasized in invariance-based approaches. This discrepancy between in-distribution accuracy and robustness under distribution shift motivates a closer examination of generalization in the presence of spurious associations, including confounding.

\section{Generalization under Confounding - unknown vs known}\label{sec:sec4}

The task of making accurate predictions on test data drawn from either the same distribution as the training data (in-distribution) or from a different distribution (out-of-distribution) is referred to as in-distribution (ID) and out-of-distribution (OOD) generalization, respectively. Age predictors often fail to adequately address the OOD challenge~\cite{luo2025bridging,cruz2024methylation,watkins2023epigenetic,min2024critical,koop2020epigenetic,tomusiak2024development}, which limits the robustness of their predictions and reduces the clinical relevance of the learned feature representations. But success of a predictive model lies in its power to predict in a new context.  Unfortunately, the problem of OOD generalization is an \emph{ill-posed} problem.  A problem is ill-posed when either of the following is applicable:
\begin{itemize}
\item Unique solution does not exist
\item Unique solution exists but computationally not feasible
\item Unique solution exists but unreliable.
\end{itemize}
In the presence of unknown confounders, OOD generalization becomes fundamentally infeasible. In particular, the  ERM principle tends to exploit spurious correlations present in the training data, which may not persist under distribution shifts. As a result, even when a unique solution exists, it may fail to generalize reliably beyond the training distribution. Therefore, we require principled approaches that enable models to generalize reliably under out-of-distribution settings.  When the set of confounders $C$ is observed, one can model the conditional mechanism $p_{Y \mid X, C}$ instead of $p_{Y \mid X}$. Under the assumption that $C$ satisfies the backdoor criterion relative to $(X,Y)$, conditioning on $C$ blocks spurious dependence induced by common causes, i.e.,
\[
Y \perp\!\!\!\perp X \mid \operatorname{do}(X), C
\]
and
\[
p_{Y \mid \operatorname{do}(X), C} = p_{Y \mid X, C}.
\]
Thus, adjusting for $C$ removes bias due to confounding and allows the conditional mechanism to better reflect the underlying causal or stable predictive relationship.
In this case, the learning problem becomes better posed, as part of the ambiguity in the $X$–$Y$ relationship can be resolved through adjustment. However, OOD generalization remains challenging: shifts in the distribution of $C$, changes in the conditional mechanism $p_{Y \mid X, C}$, or measurement noise in $C$ may still impair robustness.  A variety of strategies have been proposed to improve out-of-distribution (OOD) generalization, including domain adaptation (DA)~\cite{ben2010theory}, domain generalization (DG)~\cite{muandet2013domain,gulrajani2020search}, and distributional robust learning (DRO)~\cite{sagawa2019distributionally,kuhn2025distributionally,bai2023wasserstein}.  

\begin{tcolorbox}[colback=gray!5,colframe=gray!50]
\textbf{Takeaway}: For an age predictor to generalize reliably in OOD settings with \textbf{hidden confounding}, identifying an invariant set of features is necessary but not sufficient. Robust generalization additionally requires accounting for environment-specific variation and potential shifts in the data-generating process - a problem that can be addressed using frameworks such as the domain adaptation (DA) .
\end{tcolorbox}

We will  now discuss DA—it's underlying assumptions and inherent limitations. Most importantly, DA provides the conceptual foundation for our proposed age predictor, which we introduce in the subsequent sections.

\subsection*{Domain adaptation (DA)}

DA considers two domains: a source domain and a target domain, associated with probability distributions $\mathcal{P}^s$ and $P^t$ over the input--label space $\mathcal{X} \times \mathcal{Y}$, with $\mathcal{P}^s \neq \mathcal{P}^t$. The learner is given labeled source samples
\[
\{(x_i^s, y_i^s)\}_{i=1}^{n_s} \stackrel{\text{i.i.d.}}{\sim} \mathcal{P}^s,
\]
and unlabeled target inputs
\[
\{x_j^t\}_{j=1}^{n_t} \stackrel{\text{i.i.d.}}{\sim} \mathcal{P}^t_X,
\]
where $\mathcal{P}^t_X$ denotes the marginal of $\mathcal{P}^t$ over $\mathcal{X}$. The objective is to learn a predictor $f : \mathcal{X} \to \mathcal{Y}$ that achieves low target risk
\[
R_t(f) = \mathbb{E}_{(X,Y)\sim \mathcal{P}^t}[\ell(f(X),Y)],
\]
despite having access to labels only from the source domain.
The learned model is required to perform the \emph{same prediction task} on the target domain. In the context of age prediction, this task corresponds to estimating chronological age from molecular features. The model is expected to exploit source domain knowledge and the similarity between two domains when performing the task on the target domain. Since this is an instance of OOD generalization, the goal is to learn a \textbf{common representation} of both the domains so that we can exploit the assumption of the discriminative classifier: the training and test data comes from the same distribution. At this setting we can use the ERM principle. Rather than formalism, we are mainly interested in the assumptions of DA which will provide us practical limitations. The authors in~\cite{david2010impossibility} show that, without additional relatedness assumptions, successful domain adaptation cannot be guaranteed. One key quantity that appears in domain adaptation theory is a hypothesis-class--dependent discrepancy between the source and target \emph{marginal} input distributions, commonly defined via the $\mathcal{H}\Delta\mathcal{H}$-divergence~\citep{ben2010theory}:
\begin{equation}
\label{eq:hdeltaH_canonical}
d_{\mathcal{H}\Delta\mathcal{H}}(\mathcal{P}_X^s, \mathcal{P}_X^t)
\;=\;
2 \sup_{h,h' \in \mathcal{H}}
\left|
\mathbb{E}_{x \sim \mathcal{P}_X^s}\!\left[\mathbb{I}\{h(x) \neq h'(x)\}\right]
-
\mathbb{E}_{x \sim \mathcal{P}_X^t}\!\left[\mathbb{I}\{h(x) \neq h'(x)\}\right]
\right|.
\end{equation}

Here, $\mathcal{P}_X^s$ and $\mathcal{P}_X^t$ denote the marginal distributions over the input space $\mathcal{X}$ induced by the source and target domains, respectively. The function $\mathbb{I}\{h(x)\neq h'(x)\}$ is the indicator of disagreement between two hypotheses $h,h'\in\mathcal{H}$ on input $x$. This formulation assumes binary-valued hypotheses $h:\mathcal{X}\to\{0,1\}$ for simplicity.

Furthermore, there must be a single hypothesis (predictor) that works well (having minimum error) in both the domains. These two assumptions together are \emph{sufficient} for DA to work which is evident from the following inequality according to Theorem 2 in~\citep{ben2010theory}: 

\begin{equation}
\label{eq:da_bound}
\boxed{
\epsilon_t(h)
\;\le\;
\epsilon_s(h)
\;+\;
\frac{1}{2}\, d_{\mathcal{H}\Delta\mathcal{H}}\!\big(\mathcal{P}_X^s, \mathcal{P}_X^t\big)
\;+\;
\lambda
}
\end{equation}
where
$\lambda = \min_{h' \in \mathcal{H}} \big( \epsilon_s(h') + \epsilon_t(h') \big)$ denotes the joint error of the optimal shared hypothesis in $\mathcal{H}$.
In Equation~\ref{eq:da_bound}, the central quantity governing out-of-distribution performance is the $\mathcal{H}\Delta\mathcal{H}$ divergence. For a fixed hypothesis class $\mathcal{H}$, this term measures the maximum discrepancy in disagreement rates between pairs of hypotheses $h,h' \in \mathcal{H}$ across the source and target domains. Consequently, the divergence is small when hypotheses exhibit similar disagreement patterns on both domains. If a representation can be learned in which samples from the source and target domains become indistinguishable, then even the worst-case domain discriminator fails, leading to a reduction of the $\mathcal{H}\Delta\mathcal{H}$ divergence. Importantly, since target labels are unavailable during training, the learner cannot directly optimize target-domain error. Instead, representation learning induces a trade-off between minimizing the source-domain error and reducing the divergence term using unlabeled target data. 

\begin{tcolorbox}[colback=gray!5,colframe=gray!50]
\textbf{Takeaway}: The generalization bound for DA reveals fundamental limitations of age predictors under distributional shift. Target-domain error is governed not only by in-distribution performance, but also by the discrepancy between training and deployment environments, as measured by the $\mathcal{H}\Delta\mathcal{H}$ divergence, and by the existence of a hypothesis that generalizes across domains. In age prediction, shifts in the data-generating environment—such as changes across tissues, cohorts, sequencing protocols, or demographic groups—can induce large divergences, even when empirical risk on the training data is low.
\end{tcolorbox}

The preceding discussion naturally leads to an \textbf{adversarial learning framework}. Specifically, a feature transformation is learned to project the original inputs into a latent space in which samples from the source and target domains are indistinguishable, thereby minimizing the divergence term in Equation~\ref{eq:da_bound}. Simultaneously, a domain discriminator is trained to maximize domain separability from the same latent representation. This resulting $\min\max$ optimization defines the adversarial DA framework.
By construction, this adversarial objective suppresses features that are informative of domain identity, which are precisely the features that often give rise to spurious correlations. As such correlations encode information about \emph{where} a sample originates rather than \emph{what} is being predicted, their removal encourages reliance on domain-invariant and more stable predictive signals. 
\begin{tcolorbox}[colback=gray!5,colframe=gray!50]
\textbf{Takeaway}:  Adversarial DA is directly motivated by Equation~\ref{eq:da_bound} , as it seeks to reduce domain-informative, spurious variation by learning representations in which source and target samples are indistinguishable. While such adversarial invariance can mitigate confounding and improve robustness, genuine biological heterogeneity may still induce a large irreducible error term $\lambda$, fundamentally limiting reliable out-of-distribution generalization.
\end{tcolorbox}

\section{Confounding, Bias Mitigation, and Fairness in Chronological Age Prediction - A Closer Examination}

The preceding discussion framed domain-adversarial learning as a mechanism for suppressing domain-specific variation and promoting representations that capture predictive structure invariant across environments. In many practical learning settings, however, domain identity is often correlated with biological or demographic attributes (e.g., cohort, tissue, ancestry, or health status). Consequently, suppressing domain-informative signals does not only address robustness to distribution shift; it may also influence how learned representations depend on group-level attributes.
This observation highlights an important conceptual distinction, particularly in the context of chronological age prediction. Attributes associated with domain heterogeneity $\mathbb{S}_{exp}$ and $\mathbb{S}_{bio}$  may introduce dataset bias (see List of Definitions), correspond to $\mathbb{S}_{prot}$ relevant for fairness considerations, or—under an appropriate causal model—act as confounders influencing both the observed features and the prediction target. Although these perspectives arise from different motivations, they share a common concern: controlling how learned representations depend on auxiliary attributes.
%
While these objectives may overlap, they arise from different motivations and require separate examination. We therefore briefly discuss fairness and its relation to invariance and domain adaptation. The concept of fairness plays a crucial role in artificial intelligence (AI), particularly in the subfield of machine learning, where systems are increasingly used for decision-making. In this context, fairness generally means that a decision-making process should not rely on the set $S_{prot}$ (see List of Definitions and Mathematical Notations). In other words, decisions produced by AI systems should be independent of the set $S_{prot}$  and should not disadvantage individuals or groups based on them. There are three widely studied notions of fairness in machine learning: individual fairness, group fairness, and counterfactual fairness\cite{barocas2023fairness}.
\begin{itemize}
\item \textbf{Individual fairness} requires that similar individuals be treated similarly. That is, if two individuals are alike with respect to all task-relevant attributes, then the model’s predictions for them should be identical or differ only negligibly. For example, two individuals of the same age but different genders should receive nearly the same predicted age.
\item \textbf{Group fairness} focuses on statistical parity across predefined groups. Under this notion, different demographic groups—such as groups defined by ethnicity or gender—should, on average, receive similar outcomes. For instance, individuals of the same age belonging to different ethnic groups should have comparable predicted ages when aggregated at the group level. 
\item \textbf{counterfactual fairness} is based on causal reasoning and requires that a model’s prediction for an individual remain unchanged in a counterfactual world in which the individual’s $s \in S_{prot}$ are altered while all other relevant factors are held constant. In other words, any element of $S_{prot}$  should not causally influence the decision outcome.
\end{itemize}
Fairness is particularly relevant in age prediction tasks because the input features $X$ may be correlated $\mathbb{S}_{prot}$.  Although these attributes do not causally determine chronological age, biological and environmental differences across groups can influence molecular measurements used for prediction. Fairness considerations therefore do not require identical predictions across groups, but rather that the predictor does not exhibit systematic bias or unequal error patterns with respect to $\mathbb{S}_{prot}$.  A fairness-aware age predictor seeks representations that maintain predictive validity while avoiding consistent over- or under-estimation for specific demographic or biological subpopulations.

In our opinion, chronological age prediction from molecular profiles should explicitly consider the sample-attribute set $\mathbb{S}$ (e.g., tissue, strain, platform, protocol, sex, gender, ethnicity etc.), since variation in these attributes can induce spurious, environment-specific associations and degrade out-of-distribution generalization. The term \emph{confounder mitigation} should therefore be used cautiously in this predictive setting unless an explicit causal estimand is defined. In practice, controlling dependence on elements of $\mathbb{S}$ is more appropriately framed as \emph{bias mitigation} when the attributes belong to $\mathbb{S}_{exp}$ and/or $\mathbb{S}_{bio}$ (e.g., tissue or platform), and as \emph{fairness-aware learning} when the attributes belong to $\mathbb{S}_{prot}$ . For example, \cite{zhao2020training} describes gender as a confounder while predicting chronological age; under the classical causal definition this is not strict confounding, and Figure~5a uses double-headed arrows indicating statistical dependence rather than causal influence. A more precise interpretation is that gender can induce group-dependent statistical structure in the data, motivating invariance-based representation learning to reduce reliance on unstable or group-specific correlations rather than to correct confounding in the causal sense.

%

\textbf{So where does the confusion lie?} The conceptual confusion in so called \textbf{clock} research arises because chronological age plays two distinct roles. During model training, age functions as an observable time index, leading to a predictive task in which age causally precedes many molecular changes. In this setting, the learned mapping from molecular features $X$ to age $Y$ is fundamentally a statistical association, and classical confounding with respect to chronological age is unlikely. However, this predictive relationship is often reinterpreted biologically: clock outputs are treated as proxies for a latent biological aging state, a causal quantity that may itself be influenced by genetic, environmental, and demographic factors.  This shift in interpretation implicitly changes the causal target—from an observable time variable to an unobserved biological process—while the model was trained only on chronological age. Because the predictive direction $X \to Y$ is opposite to the plausible causal direction $Y \to X$, it becomes easy to mistake strong statistical associations for mechanistic relationships. As a result, terminology from causal inference, such as confounding, is sometimes applied to what is essentially a predictive association problem, even though the underlying causal structure differs between the training objective and the biological interpretation.

\section{Domain-Adversarial Learning in Deep Neural Networks: Modeling Complex Nonlinear Relationships}

Age-related molecular data, such as gene expression or methylation profiles, are high-dimensional and exhibit complex nonlinear structure. Simple linear models may fail to capture these patterns, motivating the use of deep neural networks for age prediction. However, deep models are particularly prone to exploiting dataset-specific signals—such as platform effects, batch artifacts, or cohort-specific biases—when trained under empirical risk minimization. As discussed earlier, such reliance on domain-specific structure leads to poor generalization under distribution shift and corresponds to a large $\mathcal{H}\Delta\mathcal{H}$-distance between domains.

To address this challenge, domain adaptation methods have been extended to deep learning frameworks through representation learning. The key idea is to learn feature representations that remain predictive for the primary task (here, age prediction) while suppressing information about domain identity. In 2015, Yaroslav Ganin and Victor Lempitsky~\cite{ganin_unsupervised_2015} proposed \textbf{Domain-Adversarial Neural Networks} (DANN), a method that integrates adversarial training into neural networks to encourage domain-invariant representations. This approach enables domain adaptation in high-dimensional settings such as transcriptomics, where nonlinear structure and complex feature interactions are prevalent. Using a technique called \emph{gradient reversal}, they showed that adversarial training of a domain classifier encourages the feature extractor to reduce the $\mathcal{H}\Delta\mathcal{H}$-distance between source and target domains (cf.\ Equation~13 in~\cite{ganin_unsupervised_2015}). Intuitively, this makes samples from different datasets harder to distinguish in the learned representation. In this sense, hiding domain information via the feature extractor is analogous to learning a \emph{fair representation} with respect to $\mathbb{S}_{prot}$ (same for mitigating bias or confounders depending on which set of attributes are considered): the model is encouraged to discard variation related to dataset identity while retaining information relevant for the prediction task.  Learning complex patterns and non-linear relationships between $\mathbb{S}$ and the set of features poorly handled by classical machine learning approaches equipped with batch effect corrections and other preprocessing techniques. Aging falls into this specific case, especially when the data are transcriptomic—high-dimensional and noisy, with complex nonlinear relationships with tissue type, cell composition, technical platform, and other biological and experimental covariates. In such settings, the aging signal is embedded within a mixture of interacting sources of variation, making it difficult to isolate using simple linear or additive models and motivating the need for representation learning approaches that can capture structured, nonlinear patterns. DANN has later been exploited to learn an invariant representation with respect to $\mathbb{S}$\cite{adeli2019representation} and also shown to be effective in practical applications of predictive modeling such as in \cite{zhao2020training}. The authors further pointed out that learning representation invariant to $\mathbb{S}$ while obtaining an optimal classifier/regressor is impossible when elements in $\mathbb{S}$  and the target variable are statistically dependent, as also pointed out in~\cite{roy2019mitigating}. In fact, the learned representation may still leak information regarding $s \in \mathbb{S}$ even after strengthening adversarial training~\cite{elazar2018adversarial}.  

\section{DANN based Chronological Age Predictor}

Building on the strategies discussed above, and guided by their theoretical soundness as well as practical applicability, we adopt these approaches in the form of a DANN based model to address age prediction while explicitly promoting bias mitigation, generalization, fairness and interpretability. Of note, the model employed for transcriptomic age prediction was originally proposed in~\cite{ferrari2022deep} by the second and the last authors of this current manuscript. In this work, we are examining it within a more explicit theoretical context, clarifying its properties and discussing its  scope and limitations for bias mitigation, fairness, OOD generalization, and interpretability. We further illustrate its practical applicability through an interventional case study (see Section~\ref{sec:sec2.4}) involving skeletal and cardiac muscle tissues, which model two distinct age-associated pathophysiological contexts: sarcopenia and cardiometabolic dysfunction, respectively. The input is the gene expression data and the output is the chronological age. The goal is to learn a \emph{debiased} and/or \emph{fair} representation with respect to $\mathbb{S}$. One important addition to our deep neural net based architecture is a \emph{Binary Stochastic Filter} (BSF)~\cite{trelin2020binary} layer at the beginning of the encoder (see Methods). The BSF mimics the $l_1$ regularization during the training phase of the neural network, helps to learn predictive genes - ensuring interpretability.

It is worth noting that most current age predictors based on transcriptomic data only partially address the issue of mitigating the effect of $\mathbb{S}$ in prediction. For instance, the BiT Age clock in \emph{C. elegans}~\cite{meyer2021bit} uses binarization to reduce noise, but does not include explicit fairness strategies. Deep learning–based models developed for human bulk transcriptomes~\cite{insilico2020patent} primarily emphasize predictive accuracy, without incorporating debiasing mechanisms. Some single-cell approaches~\cite{zakar2024profiling} apply batch correction during preprocessing, but rely on post hoc adjustments rather than integrating bias mitigation into model training. Similarly, clocks trained on brain-specific~\cite{Muralidharan2025.02.28.640749} or tissue-specific~\cite{costa2026multi} datasets limit heterogeneity by restricting the scope of the data, but do not directly address confounding during learning. In contrast, DANN explicitly includes elements of $\mathbb{S}$ in the training process. It requires minimal preprocessing (see Materials and Methods), does not assume a particular prior distribution of the data, and supports the identification of biologically relevant features in a multi-tissue setting. While adversarial learning and stochastic filtering are well-established individually, combining them into a unified framework is nontrivial. We therefore view this integration as a methodological contribution. Importantly, we present our results in a balanced manner, highlighting not only the strengths of the approach but also its limitations in light of the theoretical considerations discussed above. We believe that acknowledging both aspects is essential for the community to understand the constraints of current age predictors and to make careful claims about their scope and applicability.

\section{Results}\label{sec:sec2}
\subsection*{Adversarial Learning of $\mathbb{S}$-Invariant Representation}\label{sec:sec2.1}

\begin{figure}
    \centering
    \includegraphics[width=0.95\textwidth]{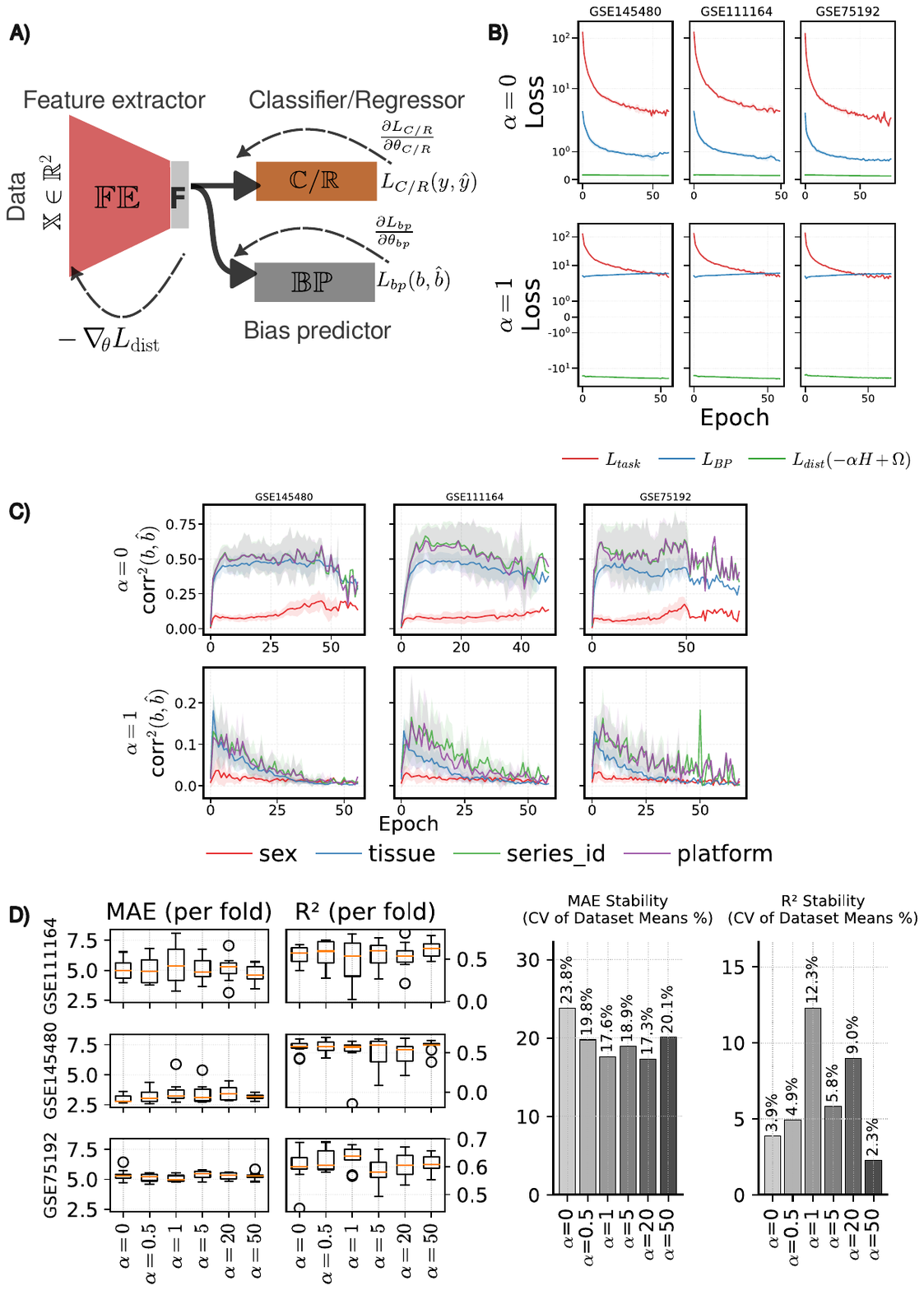}
    \caption{For detailed caption, see Section~\ref{fig:fig_captions}}
    \label{fig:figure2}
\end{figure}

In the adversarial representation learning framework described in  Section~\ref{sec:sec10_2}, the encoder ($\mathbb{FE}$) and the bias predictor ($\mathbb{BP}$) in Figure~\ref{fig:figure2}A engage in a zero-sum game.  The bias predictor ($\mathbb{BP}$) aims to infer $s\in \mathbb{S}$ from the latent representation $\mathbf{F}$, while the encoder ($\mathbb{FE}$) attempts to conceal this information resulting into a minimax optimization problem which seeks a saddle point where the representation remains predictive of age but invariant to $\mathbb{S}$. Ideally, the categorical cross-entropy loss of $\mathbb{BP}$ should decrease  during training. However, when adversarial training is active (i.e., $\alpha > 0$), the loss increases and eventually stabilizes (Figure~\ref{fig:figure2}B, second row), indicating that the encoder successfully suppresses attribute related information in the latent space. In contrast, when adversarial training is disabled ($\alpha = 0$), the loss decreases sharply, demonstrating that the protected or nuisance attributes can be readily predicted from the latent representation. This behavior is further reflected in Figure~\ref{fig:figure2}C, which shows  the correlation between predicted and true values of the attributes (one-hot encoded). The loss dynamics and correlation patterns across training epochs collectively indicate that the adversarial approach effectively mitigates bias associated with these attributes.
We further evaluate predictive performance on holdout datasets (Figure~\ref{fig:figure2}D) and assess the coefficient of variation (CV) across datasets for different values of $\alpha$. We observe that the CV of the mean absolute error (MAE) is consistently lower for $\alpha > 0$, indicating improved stability under distribution shift. This trend is less consistent for $R^2$, except at $\alpha = 50$ (see Section~\ref{sec:sec3} for further discussion).

\subsection*{Interpretable Learning via Binary Stochastic Filtering Without Significant Performance Loss}\label{sec:sec2.2}

A Binary Stochastic Filter (BSF)~\cite{trelin2020binary} is introduced at the input of the encoder to promote feature sparsification. The BSF layer operates by multiplying each input feature by a binary variable $z_j \in \{0,1\}$, where $z_j$ is sampled from a Bernoulli distribution parameterized by a learned continuous weight $w_j$. During training, the probability of retaining feature $j$ is therefore adaptively updated, allowing the model to stochastically explore different feature subsets. After training, a user-defined threshold is applied to $w_j$ to obtain a deterministic binary mask, yielding a fixed subset of selected genes for prediction (see Model architecture in Section~\ref{sec:sec10_2} for details).

\begin{figure}
         \centering
	\includegraphics[width=0.95\textwidth]{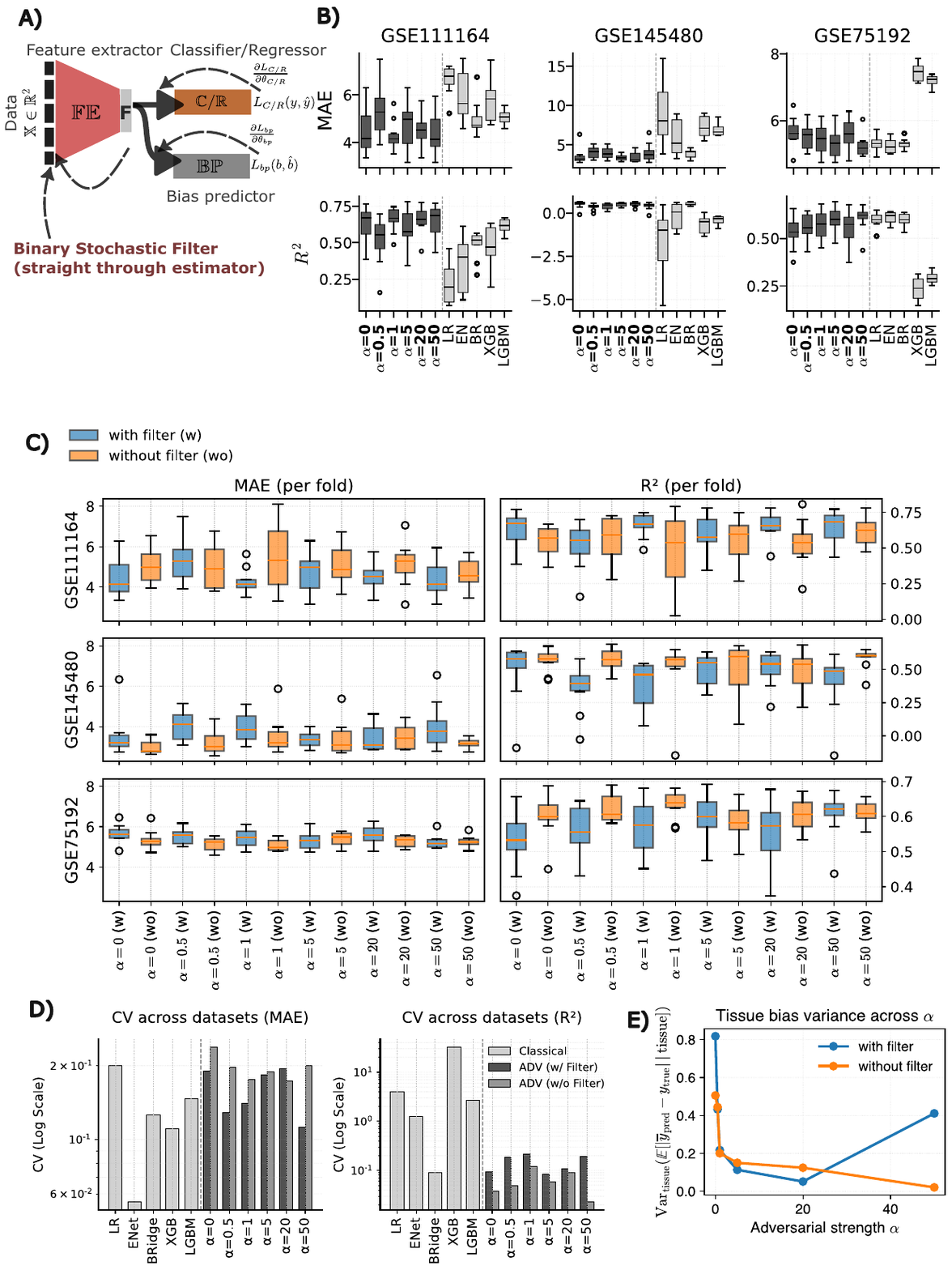}
	\caption{For detailed caption, see Section~\ref{fig:fig_captions}}	
	\label{fig:figure3}
\end{figure}

This mechanism effectively reduces the dimensionality of the input space while retaining features that contribute most to predictive performance. Empirically, incorporating the BSF layer leads to a substantial reduction in the number of genes without degrading performance and, in several cases, yields improved results compared to classical machine learning models (Figure~\ref{fig:figure3}B). When compared to the non-filtered deep model, performance remains comparable and occasionally improves (Figure~\ref{fig:figure3}C), suggesting that sparsification can mitigate overfitting in high-dimensional settings rather than merely eliminating predictive signal.

We also observe lower coefficients of variation (CV) for MAE and $R^2$ across datasets when the BSF layer is applied, indicating improved stability under dataset heterogeneity. However, CV should be interpreted with caution: a low CV does not necessarily imply superior robustness, as it depends jointly on the mean and dispersion of the metric. Therefore, improvements in stability should be evaluated alongside absolute predictive performance rather than considered in isolation.

\subsection*{Binary Stochastic Filter Selects Genes with Contextual Relevance}\label{sec:sec2.3}
\begin{figure}
         \centering
	\includegraphics[width=0.95\textwidth]{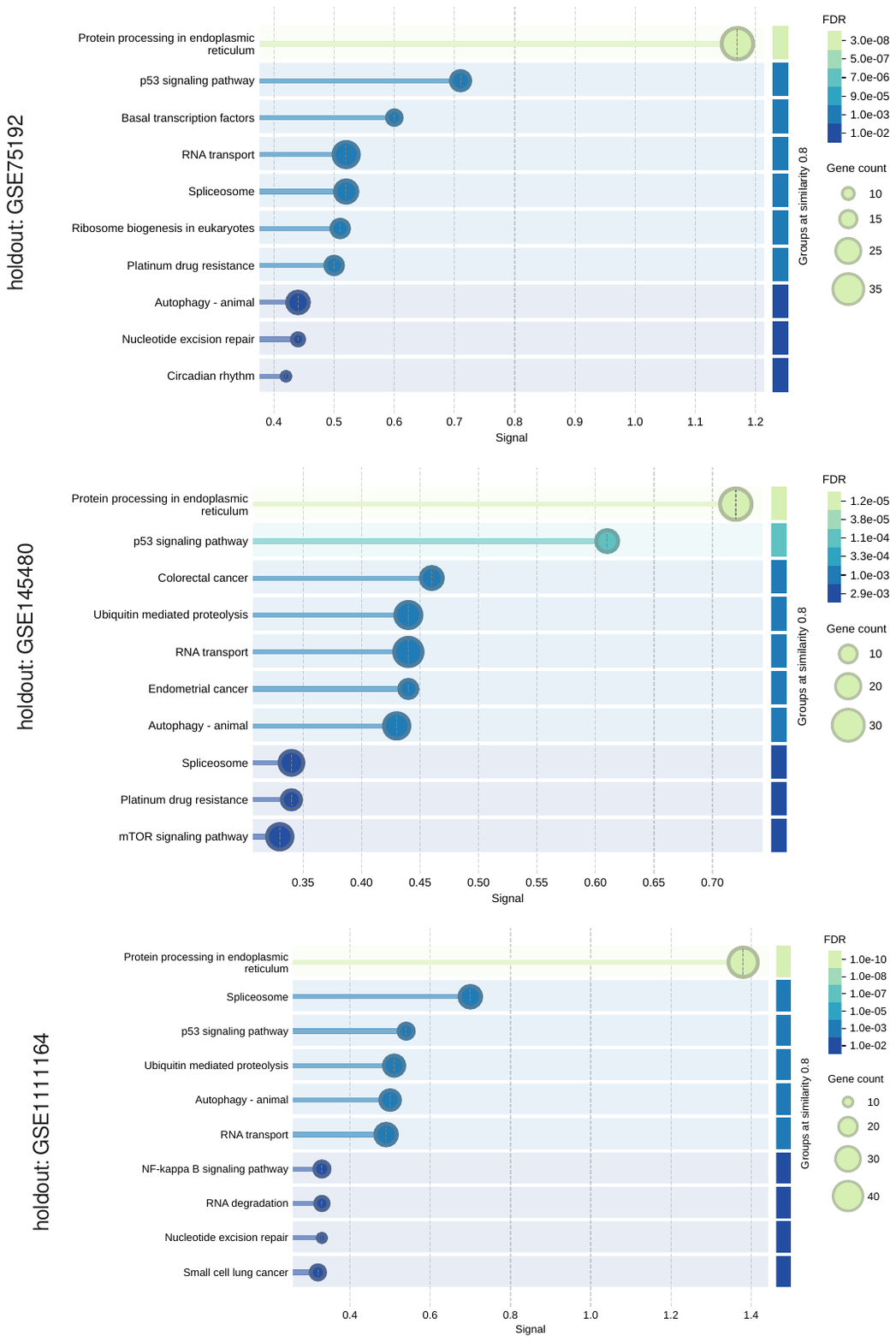}
	\caption{For detailed caption, see Section~\ref{fig:fig_captions}}	
\label{fig:figure4}
\end{figure}

For each holdout dataset, we identified the set of genes selected by the Binary Stochastic Filter in each validation fold. We then determined the intersection (common subset) of selected genes across folds to obtain a stable gene set. This consensus gene set was subsequently submitted to the STRING database (\url{https://string-db.org}) for functional enrichment analysis, and KEGG pathway enrichment was computed for the selected genes.
Across holdout settings, we noticed that the attributed genes were frequently significant and consistent with well-established aging mechanisms (see Figure~\ref{fig:figure4}). In particular, \emph{protein processing in the endoplasmic reticulum} (ER proteostasis/UPR) and \emph{autophagy} repeatedly emerged, aligning with the recognized roles of proteostasis decline and impaired macroautophagy in aging~\cite{lopez2023hallmarks,naidoo2009er,aman2021autophagy}. We also observed recurrent enrichment of the \emph{p53 signaling pathway}, consistent with extensive evidence linking DNA damage responses, cellular senescence, and p53 network activity to aging phenotypes~\cite{wu2018relevance}. Terms related to RNA handling were also prominent, including \emph{RNA transport} and, in some holdouts, \emph{spliceosome}. These results are concordant with reports that age-associated changes in RNA metabolism, including altered nuclear mRNA export and splicing dysregulation, contribute to cellular aging and age-related disease~\cite{park2022nuclear,angarola2021splicing,harries2023dysregulated}. We also noticed that, \emph{mTOR signaling} was enriched, which is consistent with the central role of nutrient-sensing pathways and mTOR in regulating aging and longevity.~\cite{saxton2017mtor,lopez2023hallmarks}. Finally, enrichment of \emph{circadian rhythm} in one case is compatible with growing evidence that circadian disruption and age-dependent clock output reprogramming are linked to aging and systemic decline~\cite{acosta2021importance,wolff2023defining}. Overall, the recurrence of these pathways across independent holdouts supports the biological plausibility of the attribution-derived gene sets and provides a pathway-level interpretation of model behavior.

\subsection*{DANN Captures Baseline Differences Across Conditions more Effectively than Conventional Regression and Tree-based Approaches}\label{sec:sec2.4}

\begin{figure}
         \centering
	\includegraphics[width=0.95\textwidth]{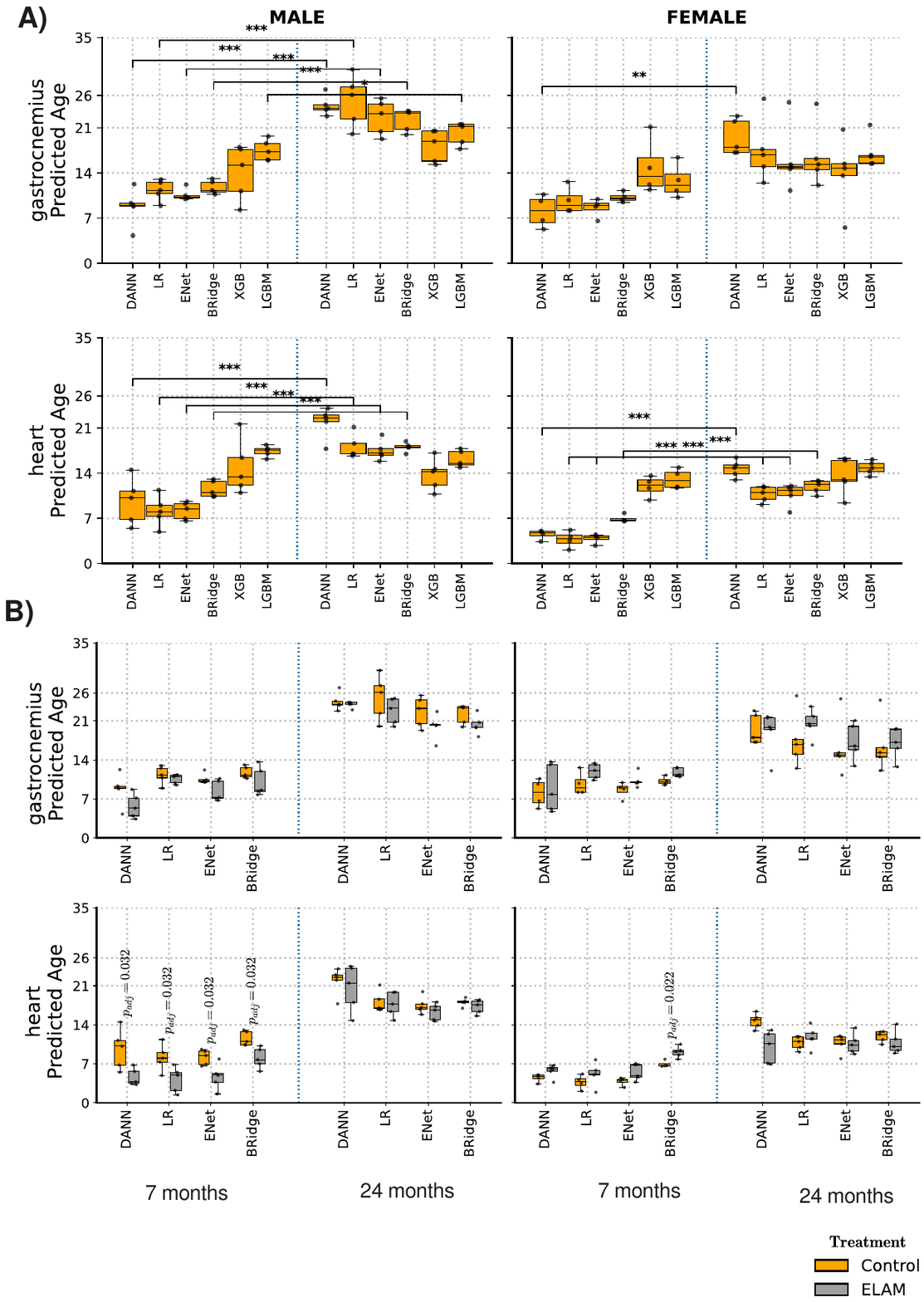}
	\caption{For detailed caption, see Section~\ref{fig:fig_captions}
}\label{fig:figure5}
\end{figure}

The ultimate test of an age-prediction model, in terms of practical applicability, is its ability to detect the effects of pharmacological interventions at the same  level of abstraction as the input data (e.g., transcriptome or epigenome). To this end, we considered a recent study by Mitchell et al.~\cite{mitchell2025mitochondria},  which investigated the effects of the ELAM peptide on skeletal and cardiac muscle in male and female mice at $7$ and $26$ months of age (the age when the samples for sequencing are collected). The authors observed that functional improvements induced by ELAM did not always translate into clear transcriptomic or epigenetic shifts, and accordingly, the chronological age predictor based on a BayesianRidge transcriptomic clock trained on a large external dataset did not strongly reflect intervention effects~\cite{tyshkovskiy2024transcriptomic}. It is important to note that the transcriptomic clock used in that study was trained on a substantially larger corpus than our own model; our dataset represents a smaller subset of the training data used in~\cite{tyshkovskiy2024transcriptomic}.
Despite this relative data scarcity, our model was able to distinguish between control groups at baseline (see Figure~\ref{fig:figure5}A) and exhibited performance that aligns with other classical predictors (see Figure~\ref{fig:figure5}B). This suggests that representation learning combined with fairness mitigation can achieve robust generalization even with limited training data — a setting that is often prone to overfitting and spurious correlations~\cite{cai2026learning}.

\section{Discussion}\label{sec:sec3}

The primary objective of this manuscript is to examine OOD generalization, bias mitigation, fairness, and interpretability as key properties of chronological age predictors—machine learning models in which chronological age is used as the target variable and is often treated as a proxy for biological age. These models are now widely applied in aging research, yet it is often unclear under what conditions such predictors generalize across heterogeneous biological settings, whether their errors differ systematically across groups (e.g., sex or experimental cohorts), and to what extent their learned structure allows meaningful biological interpretation. Furthermore, the assumptions that implicitly connect these properties to causal reasoning are rarely made explicit. By clarifying these issues, we aim to provide a more principled understanding of what chronological age predictors capture—and what they do not or can not without further assumptions. 

\subsection*{Scope of Causal Interpretations of Chronological Age Predictor} 
It is widely acknowledged that chronological age prediction is not causal in the interventional sense, the precise implications of this statement are rarely articulated. Under the invariance framework of \cite{buhlmann2020invariance}, causal interpretation arises when there exists a structural equation model in which the response variable is generated by its direct causes through an invariant mechanism across environments. In such settings, invariance can identify the causal parents of the response. However, when chronological age serves as the target variable, the structural premises required for this interpretation are biologically implausible. Chronological age represents elapsed time rather than an outcome generated by molecular features; the biologically plausible direction of causation runs from age to molecular changes, not the reverse. Consequently, even if assumptions (A1)–(A2) in Section~\ref{sec:sec2}  hold statistically and predictive relationships remain invariant across heterogeneous environments, this invariance cannot be interpreted as evidence that molecular features are causal parents of chronological age. Instead, robust age predictors should be understood as capturing stable statistical regularities induced by age-driven biological processes. Any stronger causal interpretation would require additional structural assumptions or explicit interventional validation.

\subsection*{Generalization, Confounding, Domain Adaptation, Adversarial Learning using DNN} 
In order to have a coherent narrative, we started from the perspective of learning under distributional heterogeneity. We highlighted the assumptions based on which linear predictors may approximate invariant relationships across environments and when they instead capture environment-specific or spurious associations.  This leads naturally to a discussion of generalization in the presence of latent structure, which we discussed in Section~\ref{sec:sec4}. In this context, we argue that the term \emph{confounding} must be used with caution when chronological age is the response variable. Since chronological age is not manipulable in the standard causal sense, classical confounding does not directly apply. Instead, systematic group-dependent differences in prediction errors are more appropriately framed within a fairness perspective. We therefore reinterpret certain phenomena commonly described as “confounding” as issues of fairness mitigation and provide justification for this distinction.  Among the different strategies to improve out-of-distribution (OOD) generalization, we focus on a domain adaptation (DA) framework. This choice is motivated by the fact that our final model is based on adversarial representation learning, which naturally aligns with DA principles. By encouraging domain-invariant representations, the model reduces dataset-specific effects in transcriptomic data. One may argue that domain generalization (DG) is more appropriate, since in DG the target domain is unknown during training and no adaptation is possible. This is indeed the ideal scenario, as the model should generalize to unseen domains without retraining. Although our approach is inspired by DA, its goal is consistent with DG: to learn representations that generalize across domains. Exploring dedicated DG methods in this setting is an important direction for future work.

\subsection*{Adversarial approach -  without and with Binary Stochastic Filter vs. Classical Machine Learning Models}

While adversarial regularization effectively reduces the correlation between the latent representation and nuisance attributes (Figure~\ref{fig:figure2}C), its impact on cross-dataset stability is neither uniform nor strictly monotonic. In particular, $\alpha=0$ can still yield relatively low coefficients of variation (CV) in some settings (Figures~\ref{fig:figure2}D,~\ref{fig:figure3}D, \ref{fig:figure_suppl2} and~\ref{fig:figure_suppl3}). This is not contradictory. When distributional differences across holdout datasets are limited—such as predominantly covariate shifts that do not substantially alter the conditional relationship between features and age—standard empirical risk minimization may already generalize adequately. Moreover, if dataset-specific correlations are shared across environments, exploiting them may not immediately degrade cross-dataset performance. Thus, the benefit of adversarial regularization becomes more pronounced when heterogeneity reflects mechanism shifts or group-specific structure rather than mild distributional variation. It is also important to interpret CV cautiously. Since CV is computed over dataset-level mean metrics, it is sensitive to the number of datasets and to differences in outcome variance across cohorts. This is particularly relevant for $R^2$, which depends explicitly on the variance of chronological age within each dataset and may fluctuate even when the absolute prediction error (MAE) remains stable. The more consistent reduction of CV observed for MAE therefore provides stronger evidence for improved robustness than $R^2$ alone. Taken together, these observations suggest that adversarial regularization promotes representation-level invariance, which can translate into improved cross-dataset stability when heterogeneity is driven by nuisance structure. However, the effect depends on the magnitude and nature of distributional differences, and increasing $\alpha$ does not guarantee monotonic improvement. Excessively strong adversarial constraints may suppress useful predictive signal alongside nuisance information, underscoring the need for principled tuning rather than assuming uniform benefit for all $\alpha>0$. Complementing this, the BSF layer addresses a related but distinct source of instability: high-dimensional redundancy. By inducing sparsity through $l_1$-regularization and thresholding (see Section~\ref{sec:sec10_2}), the filter restricts the model to a smaller, more informative subset of features. This reduction of representational capacity can limit overfitting to dataset-specific patterns and, when appropriately tuned, maintain or even improve predictive performance compare to the scenario when the entire set of features are used. Importantly, adversarial invariance and feature sparsification operate through different mechanisms—one suppresses group-dependent information in the latent space, while the other constrains feature complexity—yet both aim to enhance robustness under heterogeneity. Their combined effect suggests that stability across datasets is achieved not by eliminating signal, but by selectively retaining invariant and informative structure. 

It should be noted that although no automated hyperparameter optimization (e.g., Optuna~\cite{akiba2019optuna}) was performed for the adversarial framework, we were able to identify configurations that improved cross-dataset robustness relative to classical models evaluated under default \texttt{scikit-learn} settings. This observation should not be interpreted as evidence of universal superiority. Rather, it suggests that the adversarial architecture introduces an inductive bias aligned with the structured heterogeneity present in biological aging data. Classical models optimized solely for empirical risk may implicitly encode dataset-specific correlations, whereas the adversarial constraint explicitly discourages such dependence. Consequently, the observed improvements likely reflect structural alignment between the imposed invariance constraint and the data-generating characteristics, rather than an inherent advantage independent of tuning effort. A balanced hyperparameter optimization across model classes would provide a more definitive comparison.

\subsection*{Attributes are Recoverable using a Post-hoc Classifier}
The primary goal of the adversarial learning model, as described in Section~\ref{sec:sec10_2} under Materials and Methods, is to promote fairness in the learned latent representation. As shown in Figure~\ref{fig:figure2}C, the correlation between the predicted and the actual attribute value gradually decreases during training, indicating that the adversarial objective is partially effective. However, when employing a separate post-hoc probe predictor (\texttt{LogisticRegression}), we are still able to recover a considerable amount of signal corresponding to attributes from the learned representation (see Figure~\ref{fig:figure_suppl1} in the Supplementary section). For certain values of $\alpha > 0$, we observe a reduction in balanced accuracy for dataset-specific attributes, although the pattern is not strictly monotonic. On a positive note, in most cases the balanced accuracy exhibits a decreasing trend as the adversarial strength increases. Nevertheless, the signal remains detectable to a non-negligible extent.
This phenomenon is not unique to our study. Elazar and Goldberg in~\cite{elazar2018adversarial} similarly reported that a post-hoc classifier can recover a substantial amount of information from representations that were trained adversarially to suppress such attributes. Why does this occur? To address this question, let us analyze the problem starting from the theoretical framework underlying domain adaptation. Recall Equation~\eqref{eq:da_bound} that represents the standard domain-adaptation generalization bound, where,
$d_{\mathcal{H}\Delta\mathcal{H}}(\mathcal{P}_X^s,\mathcal{P}_X^t)$ quantifies how well hypotheses in the class
$\mathcal{H}$ can discriminate source from target (the $\mathcal{H}\Delta\mathcal{H}$-divergence), and
$\lambda$ is the error of the ideal joint hypothesis, i.e., the minimum achievable combined risk on both
domains, written as  $\lambda=\min_{h\in\mathcal{H}}\epsilon_s(h)+\epsilon_t(h)$.%
~\cite{ben2010theory,blitzer2007learning}. We already discussed how adversarial representation learning (e.g., gradient reversal) is directly motivated by this bound. The aim is to reduce the divergence term by learning a representation in which source and target become difficult to distinguish for a domain classifier~\cite{ganin2016domain}. However, small divergence alone does not imply successful transfer, because $\lambda$ can remain large when the labeling functions differ across domains (i.e., under conditional shift or mismatched label marginals). In that regime, enforcing invariance can conflict with maintaining low task error, and residual domain information may remain recoverable from the representation~\cite{zhao2019learning, ben2010theory}. This explains why post-hoc probing can yield substantial balanced accuracy even when the adversarial objective reduces dependence during training: adversarial training does not guarantee that all domain information is eliminated from the latent representation $Z$, especially under limited adversary capacity or when at least one $s \in \mathbb{S}$ are correlated with the prediction target. This failure mode has been documented empirically, where a separate classifier trained after adversarial training can still recover attributes from supposedly \emph{debiased} representations~\cite{elazar2018adversarial}. Importantly, in our experiments the probe accuracy decreases for attributes considered in this study when $\alpha>0$, which is consistent with partial suppression of domain signal (i.e., a reduction in the divergence-related component), even though complete removal is neither theoretically guaranteed nor empirically expected in the presence of non-negligible $\lambda$.

\subsection*{Case Study: Effects of Elamipretide (ELAM) on Skeletal and Cardiac Muscle} 
Finally, we contextualize our findings through a case study based on~\cite{mitchell2025mitochondria}, which investigates the effects of Elamipretide (ELAM), a mitochondrial-targeted peptide, in mouse models of cardiac and gastrocnemius muscle in both male and female animals. In this setting, our model not only recapitulates the primary findings reported in~\cite{mitchell2025mitochondria}, but also demonstrates a clearer separation between relevant biological groups. Importantly, the model is able to distinguish between control groups across biological strata with high statistical confidence, whereas classical machine learning models with default configurations fail to do so in case of gastrocnemius muscle for Female. This distinction is critical: if a predictor cannot reliably resolve baseline differences between control groups, then any observed shift under intervention becomes difficult to interpret. In other words, detecting treatment effects presupposes that the model captures the underlying structure of the control condition with sufficient resolution and stability. Without this baseline discriminative capacity, apparent intervention effects may reflect noise, model instability, or uncontrolled heterogeneity rather than genuine biological response. 

\subsection*{Causal Biomarkers of Aging} 
In a recent work by Ying et al.~\cite{ying2024causality} on so-called \emph{causal biomarkers} in epigenetic aging, the authors used large-scale genetic data and epigenome-wide Mendelian randomization (EWMR) to identify CpG sites that are putatively causal for several aging-related traits, and showed that existing epigenetic clocks are not enriched for these causal CpGs. Based on this causal CpG set, they constructed causality-enriched clocks with improved performance compared to conventional clocks trained on all methylation sites. While this strategy incorporates external causal information, it is not fully data-driven in the sense of discovering causal features directly from methylation data alone; instead, the causal candidate set is preselected using genetic instruments and then fed into standard \emph{clock} models.  

\section{Conclusions}

Based on our findings, we suggest a careful consideration of generalization behavior, fairness constraints, interpretability, and underlying structural assumptions before deploying such models in experimental or translational contexts. Robust predictive performance alone does not justify causal interpretation. At the same time, developing a fully data-driven framework that simultaneously enforces fairness, promotes sparsity, and provides identifiable causal guarantees remains an open challenge. Bridging representation-level invariance, high-dimensional feature selection, and formal causal reasoning in complex biological systems represents an important and promising direction for future research.

\section{Materials and Methods}\label{sec:sec10}

All experiments consisted of training neural network models implemented in Python 3.6.8 (\url{https://www.python.org/downloads/release/python-368/}) with Tensorflow 1.13.1 (\url{https://pypi.org/project/tensorflow/1.13.1/}) on Ubuntu 22.04. Figures were assembled using Inkscape version 1.4 (\url{https://inkscape.org}) and exported as PDF.

\subsection*{Bulk RNA-sequencing dataset}\label{sec:sec4.1}
 We consider six publicly available bulk RNA-seq datasets: 
\href{https://www.ncbi.nlm.nih.gov/geo/query/acc.cgi?acc=GSE132040}{GSE132040}, 
\href{https://www.ncbi.nlm.nih.gov/geo/query/acc.cgi?acc=GSE141252}{GSE141252}, 
\href{https://www.ncbi.nlm.nih.gov/geo/query/acc.cgi?acc=GSE111164}{GSE111164}, 
\href{https://www.ncbi.nlm.nih.gov/geo/query/acc.cgi?acc=GSE145480}{GSE145480}, 
\href{https://www.ncbi.nlm.nih.gov/geo/query/acc.cgi?acc=GSE75192}{GSE75192}, and the dataset from 
\href{https://www.ncbi.nlm.nih.gov/geo/query/acc.cgi?acc=GSE280699}{GSE280699} from~\cite{mitchell2025mitochondria} as a case study (see Figure~\ref{fig:figure5}).

To facilitate cross-species analysis in the future, we restrict our feature space to the set of one-to-one orthologous genes between humans and killifish with at least $70\%$ sequence similarity. This enables downstream investigation of how the gene set identified by the BSF layer relates to conserved aging processes across species. Our goal is to identify candidate aging biomarkers that are potentially species-invariant.

\subsection*{Preprocessing}\label{sec:sec10_1}

For each dataset, we employ minimal preprocessing. For each dataset, genes with zero total counts are filtered out followed by exclusion of genes having length shorter than $500$bp (according to GENCODEvM19 annotations). After that, samples are grouped by tissue type and only tissues with with at least a minimum number of samples considered. The minimum number of samples is set to $\max\!\left(1,\;\left\lceil 0.2\, N_{\text{tissue}} \right\rceil\right)$, where $N_{\text{tissue}}$ is the number of samples in that tissue. A gene was then classified as expressed in that tissue if the count was at least 10 in those samples ($\geq 20 \%$ of tissue-specific samples). The common genes are then considered to be the final gene set for that particular dataset. These thresholds can be adjusted depending on the characteristics of the dataset.

The count matrix is first transformed to counts per million (CPM), followed by a $\log_2(c+1)$ transformation, where $c$ denotes the CPM-transformed count. This helps reduce the effect of sequencing depth differences and stabilizes variance across genes.  After these transformations, features are standardized using the mean and standard deviation computed from the training data. The same statistics are then applied to the test data. This ensures that scaling is consistent across splits and prevents information leakage from the test set into the training process.

\subsection*{Model Architecture}\label{sec:sec10_2}

Our model combined the architecture from~\cite{adeli2019representation} and a binary stochastic filter from~\cite{trelin2020binary}. The loss function is same as that of (cf.\ Equation~3 in~\cite{adeli2019representation}), with the exception is that for the bias predictor and distiller loss we consider categorical cross-entropy. Detailed model summary is given by Table~\ref{tab:model_architecture}.
\begin{table}[ht]
\centering
\small
\begin{tabular}{lll}
\hline
\textbf{Module} & \textbf{Layer} & \textbf{Dimension / Description} \\
\hline
\multicolumn{3}{c}{{Feature Encoder: $\mathbb{FE}: X \rightarrow$ \textbf{F}}} \\
\hline
Input & Input layer & $(d)$ features \\
 & Binary Stochastic Filter & Bernoulli mask per feature (learned prob.) \\
Block 1 & Dense + BN + ReLU & $d \rightarrow 256$ \\
Block 2 & Dense + BN + ReLU + Dropout & $256 \rightarrow 256$ \\
Block 3 & Dense + BN + ReLU + Dropout & $256 \rightarrow 106$ \\
Regularization & Gaussian Noise (optional) & std = 0.05 \\
Bottleneck & Dense + BN + ReLU & $106 \rightarrow 64$ \\
Output & Linear projection & $64 \rightarrow \dim(\textbf{F})$ \\
\hline
\multicolumn{3}{c}{{Target Predictor: $\mathbb{C/R}: \textbf{F} \rightarrow y$}} \\
\hline
Hidden 1 & Dense + ReLU + L2 & $\dim(\textbf{F}) \rightarrow \frac{\dim(\textbf{F})}{2}$ \\
Hidden 2 & Dense + ReLU + L2 & $\frac{\dim(\textbf{F})}{2} \rightarrow \frac{\dim(\textbf{F})}{4}$ \\
Output & Linear & $\rightarrow 1$ (regression) \\
\hline
\multicolumn{3}{c}{{Bias Predictor (Adversary): $\mathbb{BP}: \textbf{F} \rightarrow \mathbb{S}$}} \\
\hline
Trunk & Dense + BN + ReLU + Dropout & $\dim(\textbf{F}) \rightarrow 256$ \\
 & (repeated trunk\_depth times) &  \\
Head(s) & Dense + ReLU + Dropout & $256 \rightarrow 128$ (per confounder) \\
Output & Linear / Softmax & per confounder class count \\
\hline
\end{tabular}
\caption{Summary of the encoder, predictor, and adversarial bias network architecture as depicted in Figure~\ref{fig:figure2}A.}
\label{tab:model_architecture}
\end{table}
In our model architecture, we introduce batch normalization that stabilizes training and reduces internal covariate shift~\cite{ioffe2015batch}, while dropout introduces stochastic regularization that improves generalization by preventing co-adaptation of features~\cite{srivastava2014dropout}. The Gaussian noise layer further regularizes the representation by injecting small perturbations, a technique known to enhance robustness and smooth decision boundaries~\cite{bishop1995training}. The inclusion of a pre-\textbf{F} bottleneck enforces dimensional compression, consistent with the information bottleneck principle~\cite{tishby2000information}, encouraging the representation to retain task-relevant information while discarding nuisance variability. The target predictor employs linear output activation appropriate for continuous regression targets, with $l_2$ regularization (weight decay) to control model complexity~\cite{krogh1991simple}.  The adversarial bias predictor is structured as a high-capacity multi-head network. The shared trunk extracts confounder-relevant structure, while per-head towers allow modeling of heterogeneous attributes (e.g., tissue, batch, sex). The adversarial setup follows the minimax principle used in domain-adversarial learning~\cite{ganin2016domain}, where the encoder learns representations predictive of the target while minimizing information regarding attributes in $\mathbb{S}$.

\subsubsection*{Feature Selection using a Regularized BSF}
The feature selection unit of our model consists of two components: a Binary Stochastic Filter adopted from \cite{trelin2020binary} and a $l_1$ regularizer. 
The BSF layer acts as a trainable feature gate that learns, for each gene, whether it should be retained or discarded. By randomly switching genes on and off during training, the model can evaluate their contribution to predictive performance. As learning progresses, informative genes converge to high keep-probabilities, while uninformative ones are driven toward zero, yielding an automatically discovered subset of relevant features. Complementing this mechanism, the $l_1$ regularizer imposes an effective upper bound on the number of genes the model can rely on. This penalty discourages reliance on too many features simultaneously and steers the network toward a compact, parsimonious representation composed of the most informative genes. Notably, the BSF operates similarly to Dropout in that it randomly deactivates input features during training. However, unlike standard dropout, where dropout probabilities are fixed and uniform, the filter learns feature-specific dropout probabilities \( w_i \) through adversarial training. This adaptivity enables the network to selectively suppress features correlated with the protected variable \( B \), thereby achieving targeted de-confounding rather than generic regularization.

\subsection*{Model Training}\label{sec:sec10_3}

For model development, we split the data into a training set and independent holdout sets. We employ a leave-one-set-out (LOSO) validation strategy, in which GSE111164, GSE145480, and GSE75192 are each treated as holdout datasets, while the remaining datasets are used for training. In an alternative setup, we train exclusively on GSE132040 and GSE141252 and evaluate the model on GSE111164, GSE145480, and GSE75192. We deliberately retain GSE132040 and GSE141252 as training datasets in both setups due to their larger sample sizes and diverse multi-tissue composition, which provide a broad representation of biological and technical variability for model learning. To avoid transductive data leakage (information from the test set unintentionally influences the training process), gene selection is performed using only the training data. Genes not present in the training set but appearing in holdout datasets are set to zero during evaluation. This ensures that the structure or distribution of the holdout data does not influence model training, preserving a strict separation between training and evaluation domains.

For each cross-validation fold, the model is trained using an early stopping criterion to prevent overfitting. Training proceeds in epochs, where each epoch consists of $50$ stochastic gradient steps. At every step, we sample a mini-batch of size $64$ from the training split. Each mini-batch contains the gene expression input, the target age, and the attributes: sex, tissue, platform, and series ID which are one-hot encoded.

For the Elamipretide intervention study, we did not employ a cross-validation strategy. Instead, we trained the model on the combined dataset consisting of GSE145480, GSE75192, GSE132040, and GSE141252 to maximize the available training data and improve model generalization across studies. The model was trained for 500 epochs, with checkpoints saved every 10 epochs to monitor performance throughout training. The best performance was achieved at epoch $180$, which corresponds to the result presented in Figure~\ref{fig:figure5}. The batch size for this case is set to $128$. Training alternates between three coupled updates that balance prediction performance and bias mitigation:
\begin{enumerate}
\item \textbf{Bias predictor ($\mathbb{BP}$) update}  
In this step, $\mathbb{FE}$ is kept fixed and only the $\mathbb{BP}$  is updated. The $\mathbb{BP}$ attempts to classify attributes from the current latent representation. We run five updates of this component per training step so that the $\mathbb{BP}$ becomes a strong \emph{attacker}, meaning it learns to extract as much information about $s \in \mathbb{S}$ as possible from the representation. This is important because a weak $\mathbb{BP}$ would fail to expose residual information of attributes in the latent space.

\item \textbf{Adversarial Representation Update}  
In this phase, the $\mathbb{BP}$ is frozen and only the $\mathbb{FE}$ is updated. The goal is to modify the latent representation so that the $\mathbb{BP}$ becomes less able to recover attribute signal. This is implemented using an adversarial loss based on the cross-entropy of the $\mathbb{BP}$'s outputs, scaled by the hyper-parameter $\alpha$. The gradient from this loss is reversed before being passed to the $\mathbb{FE}$, which encourages the $\mathbb{FE}$ to suppress attribute-related information in the representation. We perform two such updates per training step. Intuitively, while the bias predictor learns to detect attributes, the $\mathbb{FE}$ learns to hide them.

 \item \textbf{Age Predictor Update}  
Finally, the model is updated to improve age prediction. This step ensures that, while information of attributes is being suppressed, the representation still retains meaningful biological signal related to aging.
\end{enumerate}

Together, these alternating updates approximate a minimax-style training process: the $\mathbb{BP}$ continuously tries to recover information about attributes, while the $\mathbb{FE}$  learns to prevent this, and at the same time supports accurate age prediction. The strength of the adversarial signal is controlled by the hyper-parameter $\alpha$. We include a burn-in period of $50$ epochs before applying early stopping. During early training, the adversarial interaction between the $\mathbb{FE}$  and the $\mathbb{BP}$ can be unstable, and validation performance may fluctuate. Allowing this initial phase gives the model time to reach a more stable regime before model selection is applied. For each fold, the model checkpoint with the lowest validation error in age prediction is selected. For this best-performing model, we store: The trained model weights, The latent representations for both training and validation data (used for post-hoc probing of attribute leakage), and a ranked list of selected genes derived from the BSF layer.

To obtain the gene list, we extract the BSF layer weights and retain genes whose weight exceeds a fixed threshold (e.g., $0.5$). These genes are sorted by weight, producing an interpretable set of candidate biomarkers that contribute most strongly to the learned aging representation.

\subsection*{Model Hyperparameters}\label{sec:sec10_4}
The BSF layer includes a sparsity regularizer with strength $10^{-2}$ and a cut threshold of $3000$. In simple terms, the regularizer looks at the \emph{sum} of all BSF gene weights. As long as this total stays below $3000$, no penalty is applied. Once the total weight exceeds this threshold, a penalty is introduced that increases linearly with the excess amount. This encourages the model to keep the overall number (or total magnitude) of selected genes under control, while still allowing flexibility early in training. In other words, the model is not immediately forced to be sparse, but is gradually pushed to reduce the number of active genes if too many become important.

For the leave-one-set-out experiments (with/without filter, as shown in Figures~\ref{fig:figure2} and~\ref{fig:figure3}), the dimension of the encoded feature space is set to $60$.  For the Elamipretide intervention study as shown in Figure~\ref{fig:figure5}, the dimension is set to $40$.  For both the cases, the learning rates are set to $3\times 10^{-4}$, $2\times 10^{-4}$, and $10^{-3}$ for the bias predictor, the distiller (adversarial encoder update), and the age predictor respectively. The age predictor is trained with the largest learning rate so it can quickly learn the main task and stabilize the representation. The adversarial components use smaller learning rates because they operate in a competitive (minimax) setting: the bias predictor tries to detect confounders while the encoder tries to hide them. If these parts are updated too aggressively, training can become unstable and oscillatory. Using more conservative learning rates helps maintain stable training while still allowing the model to reduce attribute information over time.

\section{Figure Captions}\label{fig:fig_captions}

\noindent\textbf{Figure~\ref{fig:figure2}.}{\textbf{Working principle of adversarial age predictor.} \textbf{A)} The model consists of a feature extractor that projects high-dimensional input features into a latent representation $\mathbf{F}$. An age regressor predicts chronological age from $\mathbf{F}$, while a bias predictor attempts to infer attributes in $\mathbb{S}$, which are sex, tissue, series ID and platform, from the same latent space. The feature extractor and bias predictor are trained in an adversarial zero-sum game: the bias predictor minimizes the categorical cross-entropy loss to correctly classify the attributes in $\mathbb{S}$, whereas the feature extractor is optimized using a distiller loss defined as the negative categorical cross-entropy (see Section~\ref{sec:sec10_3}). \textbf{B)} For each holdout dataset (indicated in the title), loss dynamics are shown for $\alpha = 0$ (no adversarial component) and $\alpha = 1$, representing the adversarial component turned off and on, respectively. The plotted losses include $L_{\mathrm{task}}$ (age regression loss), $L_{\mathrm{BP}}$ (bias predictor loss), and $L_{\mathrm{dist}}$ (distiller loss). The distiller loss is defined as $L_{\mathrm{dist}} = -\alpha H + \Omega$, where $H$ denotes the categorical cross-entropy of the bias predictor, and $\Omega$ represents the regularization term imposed by Keras (corresponding to the weighting of loss components associated with attributes in $\mathbb{S}$). Increasing $\alpha$ strengthens the adversarial pressure applied to the latent representation. \textbf{C)}  Average squared correlation ($r^2$) between the predicted $s\in \mathbb{S}$ and its true value on the training data, shown for each holdout dataset at $\alpha=0$  and $\alpha=1$. For categorical attributes, labels are encoded in one-hot form and $r^2$ is computed per class and averaged across classes. In this setting, correlation between binary (0/1) indicators and predicted probabilities is equivalent to the point-biserial correlation, providing a continuous measure of linear dependence between the true attribute and model predictions.  \textbf{D)} \emph{Left}. Prediction performance per fold in terms of MAE (Mean Absolute Error) and $R^2$ (coefficient of determination) for range of $\alpha=[0,0.5,1,5,20,50]$ varying the strength of adversary.  \emph{Right}. Cross-dataset stability of the adversarial neural network as a function of adversarial strength $\alpha$.  For each $\alpha \in \{0, 0.5, 1, 5, 20, 50\}$, MAE and $R^2$ were computed independently for each fold within each dataset and then averaged across folds to obtain a single mean performance value per dataset. The coefficient of variation (CV) was subsequently calculated across the three dataset-level mean values as $ \mathrm{CV} = 100 \times \frac{\sigma(\bar{M}_1,\bar{M}_2,\bar{M}_3)}
{\mu(\bar{M}_1,\bar{M}_2,\bar{M}_3)}$, where $\bar{M}_d$ denotes the fold-averaged performance for dataset $d$, and $\mu$ and $\sigma$ represent the mean and standard deviation across datasets, respectively.}

\vspace{10pt}

\noindent\textbf{Figure~\ref{fig:figure3}.}{\textbf{Adversarial approach with Binary Stochastic Filter.}
\textbf{A)} A Binary Stochastic Filter is introduced as the first layer of the encoder module. 
\textbf{B)} Performance comparison between classical and adversarial approaches across the indicated holdout datasets. 
\textbf{C)} Boxplots of MAE and $R^2$ across folds for varying adversarial strengths $\alpha$. 
\textbf{D)} Coefficient of variation (log scale) computed following the same procedure as in Figure~\ref{fig:figure2}D\emph{right}, additionally including the adversarial model with the stochastic filter.
\textbf{E)} Tissue bias as a function of adversarial strength for GSE75192 (as it has multiple tissues). Bias is quantified as the variance of the mean absolute residual across tissues,$\mathrm{Var}_t(\mathbb{E}[|\overline{y}_{\mathrm{pred}}-y_{\mathrm{true}}| \mid t])$. Lower values correspond to more homogeneous prediction errors across tissues.
Increasing adversarial strength initially reduces tissue-dependent bias, but excessive adversarial pressure can increase variability by removing informative biological signal. A detailed residual plot is given by Figure~\ref{fig:figure_suppl4}
}
\vspace{10pt}

\noindent\textbf{Figure~\ref{fig:figure4}.}\textbf{KEGG pathway enrichment analysis using STRING.}
Gene sets were derived from the DANN model trained with adversarial strength $\alpha = 50$, which controls the contribution of the domain-adversarial loss during training. Only genes consistently identified across all cross-validation folds for a given holdout dataset were retained to ensure robustness. Protein--protein interaction networks were constructed using the STRING database (\url{https://string-db.org}) considering only high-confidence interactions (minimum interaction score = 0.7). Active interaction sources were restricted to experimentally validated interactions and co-expression evidence to increase biological specificity. KEGG pathway enrichment was performed using the whole genome as background. Pathways were considered significant at a false discovery rate (FDR) $\leq 0.05$ after multiple-testing correction. Additional filtering criteria included enrichment signal $\geq 0.01$ and strength $\geq 0.01$, where strength reflects the log$_{10}$ ratio of observed to expected gene counts within a pathway. Enriched terms were clustered based on similarity (threshold $\geq 0.8$) to reduce redundancy and highlight coherent biological themes. The results corresponds to $\alpha = 50$.

\vspace{10pt}

\noindent\textbf{Figure~\ref{fig:figure5}.}\textbf{Elamipretide intervention: a case study from~\cite{mitchell2025mitochondria}.}
Predicted age from DANN and classical regression models (LR,  ENet,  BRidge, XGB, LGBM) in gastrocnemius and heart tissues, stratified by sex (male, female) and chronological age (7 vs.\ 26 months). Models were trained ($\alpha=1$) on the combined dataset consists of \href{https://www.ncbi.nlm.nih.gov/geo/query/acc.cgi?acc=GSE132040}{GSE132040}, 
\href{https://www.ncbi.nlm.nih.gov/geo/query/acc.cgi?acc=GSE141252}{GSE141252}, 
\href{https://www.ncbi.nlm.nih.gov/geo/query/acc.cgi?acc=GSE145480}{GSE145480}, 
\href{https://www.ncbi.nlm.nih.gov/geo/query/acc.cgi?acc=GSE75192}{GSE75192},  and evaluated on the the dataset \href{https://www.ncbi.nlm.nih.gov/geo/query/acc.cgi?acc=GSE280699}{GSE280699} from~\cite{mitchell2025mitochondria} . Each point represents one biological sample; boxes indicate the median and interquartile range (IQR), with whiskers extending to 1.5$\times$ IQR. \textbf{A)} Age-associated differences within control animals (7 vs.\ 26 months). Horizontal brackets denote two-sided Welch's $t$-tests comparing age groups within each model. \textbf{B)} Treatment-associated differences (ELAM vs.\ Control) within each tissue $\times$ sex $\times$ age stratum for each model. For both panels, $p$-values were adjusted for multiple testing using the Benjamini--Hochberg false discovery rate (FDR) within each tissue $\times$ sex comparison across models, yielding adjusted values ($p_{\mathrm{adj}}$). Significance thresholds: $^{***}p_{\mathrm{adj}} < 0.001$, $^{**}p_{\mathrm{adj}} < 0.01$, $^{*}p_{\mathrm{adj}} < 0.05$, ns: not significant ($p_{\mathrm{adj}} \geq 0.05$). Only significant comparisons are annotated.

\section*{Data and Code availability}

\subsection*{Data}
This paper analyzes existing, publicly available data, accessible under the corresponding GEO terms, \href{https://www.ncbi.nlm.nih.gov/geo/query/acc.cgi?acc=GSE132040}{GSE132040}, 
\href{https://www.ncbi.nlm.nih.gov/geo/query/acc.cgi?acc=GSE141252}{GSE141252}, 
\href{https://www.ncbi.nlm.nih.gov/geo/query/acc.cgi?acc=GSE111164}{GSE111164}, 
\href{https://www.ncbi.nlm.nih.gov/geo/query/acc.cgi?acc=GSE145480}{GSE145480}, 
\href{https://www.ncbi.nlm.nih.gov/geo/query/acc.cgi?acc=GSE75192}{GSE75192}, and the dataset from 
\href{https://www.ncbi.nlm.nih.gov/geo/query/acc.cgi?acc=GSE280699}{GSE280699}. Additional files are deposited to Zenodo and are publicly available as of the date of peer-reviewed publication.
\subsection*{Code}
All original code has been deposited at Zenodo and is publicly available as of the date of peer-reviewed publication.

\section*{Acknowledgements}
The authors would like to thank Leibniz Institute on Aging - Fritz Lipmann Institute (FLI), Jena, Germany  and the Core Facilities and Services  of the FLI  for their technological  and infrastructural support. The FLI is a member of the Leibniz Association and is financially supported by the Federal Government of Germany and the State of Thuringia. 
\subsection*{Funding Statement}
This study was supported by the German Research Foundation (DFG) (Grant number: 830041) and the Federal Ministry of Research, Technology and Space of Germany (BMFTR) via the GoBio initial program 2024-25 (Grant number: FKZ 03LW0596). The funding agencies did not influence the design of the study, the collection, analysis, and interpretation of data, nor the manuscript writing. The responsibility for the content of this publication lies with the authors.

\section*{Declaration of generative AI and AI-assisted technologies in the writing process}
During the preparation of this work the author(s) used ChatGPT (OpenAI, GPT-5.2) in order to improve the readability and language of the manuscript. After using this tool/service, the author(s) reviewed and edited the content as needed and take(s) full responsibility for the content of the published article.

\section*{Author Contributions}
\begin{itemize}
\item Conceptualisation: DP, EF, AC
\item Writing – Original Draft Preparation: DP
\item Computations and Preparation of figures: DP
\item Funding Acquisition: AC, DP
\item Review and Editing: DP, EF, IG, AC. 
\end{itemize}


\section*{Declaration of Interests}

Elisa Ferrari and Alessandro Cellerino are co-authors of a submitted patent (\href{https://worldwide.espacenet.com/patent/search/family/082655247/publication/WO2024017780A1?q=WO2024017780A1}{WO2024017780A1}) for lifespan and healthspan prediction based on the adversarial approach described here.

\section*{Supplementary information}\label{sec:suppl}
\setcounter{figure}{0}
\renewcommand{\thefigure}{SA\arabic{figure}}
\begin{figure}[H]
         \centering
	\includegraphics[width=\textwidth]{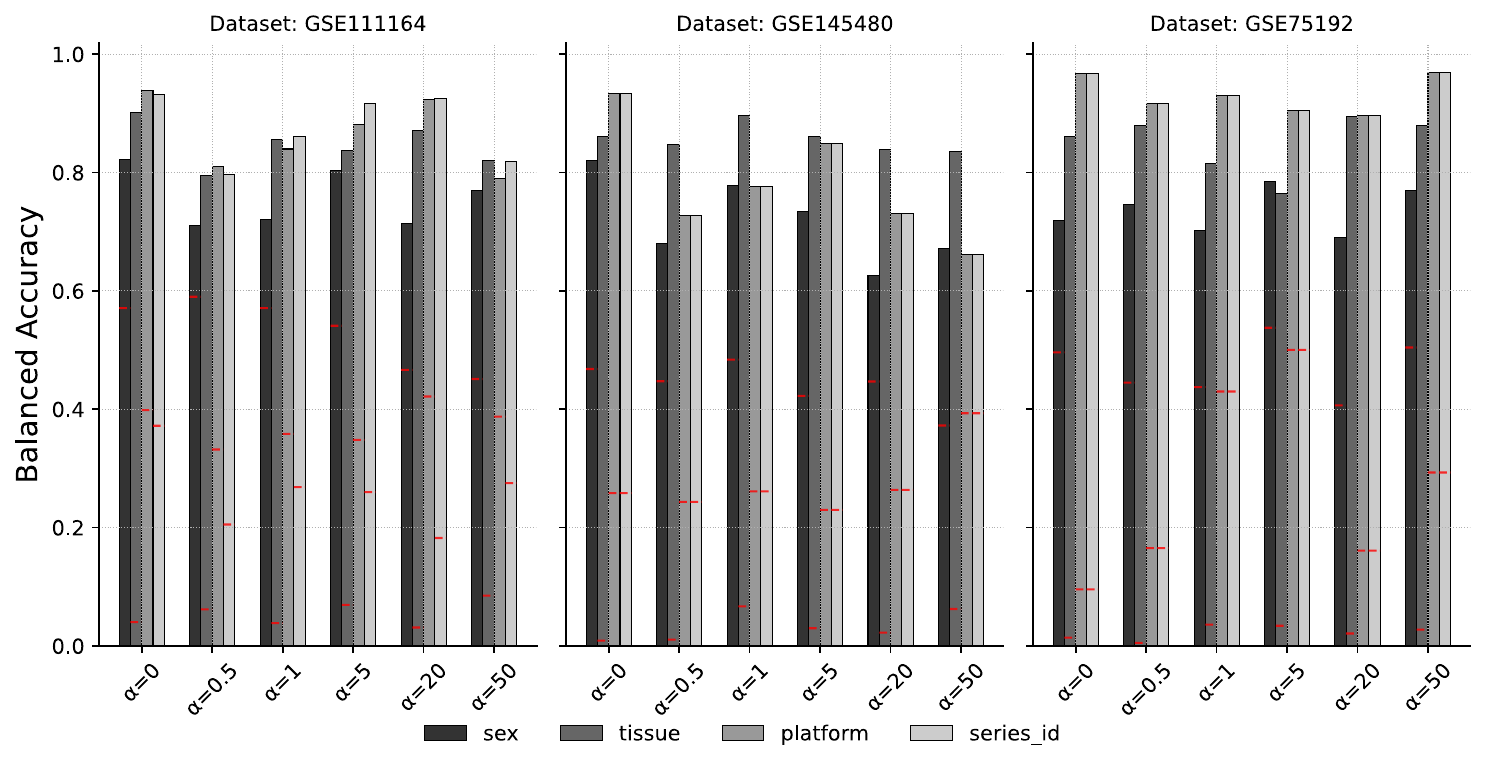}
	\caption{
\textbf{Recovery of attributes in $\mathbb{S}$ using Post-hoc Analysis}
Balanced accuracy of a linear probe (\texttt{LogisticRegression}) trained on the learned latent representation $Z$ to predict attributes: sex, tissue, platform, and series ID, for each holdout dataset. Bars correspond to probe performance for different adversarial strengths $\alpha$. Red dashed lines indicate the permutation baseline, obtained by training the probe after randomly shuffling the attribute labels, thereby representing chance-level performance.  A decreasing balanced accuracy with increasing $\alpha$ suggests partial suppression of confounder-related information in $Z$. However, residual predictability remains, indicating that attribute signals are not fully eliminated from the representation. This observation is consistent with prior work showing that adversarial objectives do not guarantee complete removal of attribute information, and that post-hoc classifiers may still recover such signals from \emph{debiased} representations~\cite{elazar2018adversarial} (see Discussions).
}	
	\label{fig:figure_suppl1}
\end{figure}

\begin{figure}[H]
         \centering
	\includegraphics[width=\textwidth]{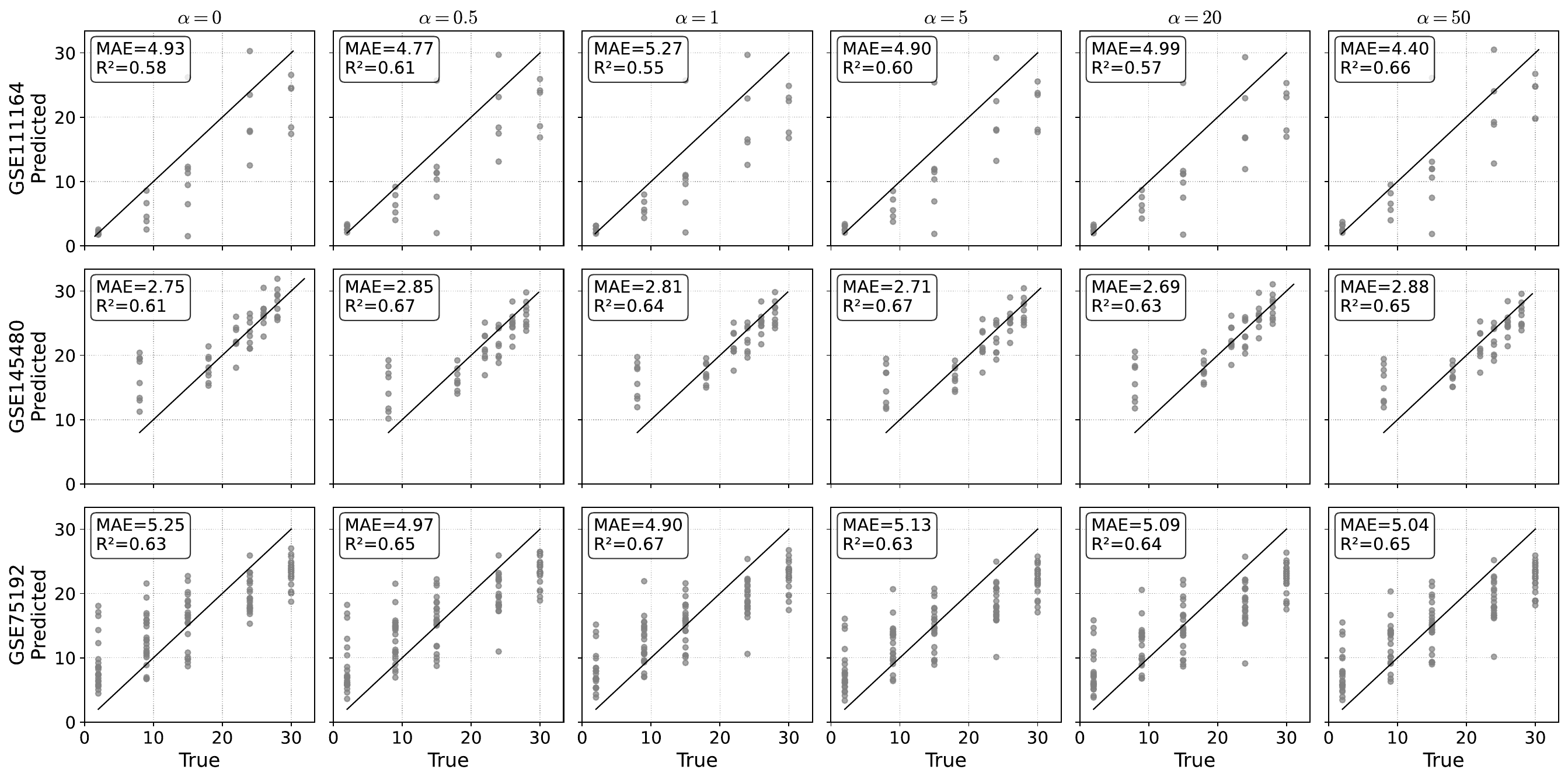}
	\caption{\textbf{Regression plots of fold-averaged predictions \emph{without the filtering layer} (Figure~\ref{fig:figure2}A) for each leave-one-dataset-out (LOSO) holdout setup.}
Each row corresponds to a different holdout dataset (GSE111164, GSE145480, GSE75192), and each column represents a different domain-adversarial strength ($\alpha$). 
For every subplot, predictions from all LOSO folds were aggregated and averaged per biological sample. 
Points represent the mean predicted age versus chronological age. 
Performance metrics (MAE and $R^2$) are computed using the averaged predictions. 
The diagonal line indicates the identity line ($y=x$).
}\label{fig:figure_suppl2}
\end{figure}

\begin{figure}[H]
         \centering
	\includegraphics[width=\textwidth]{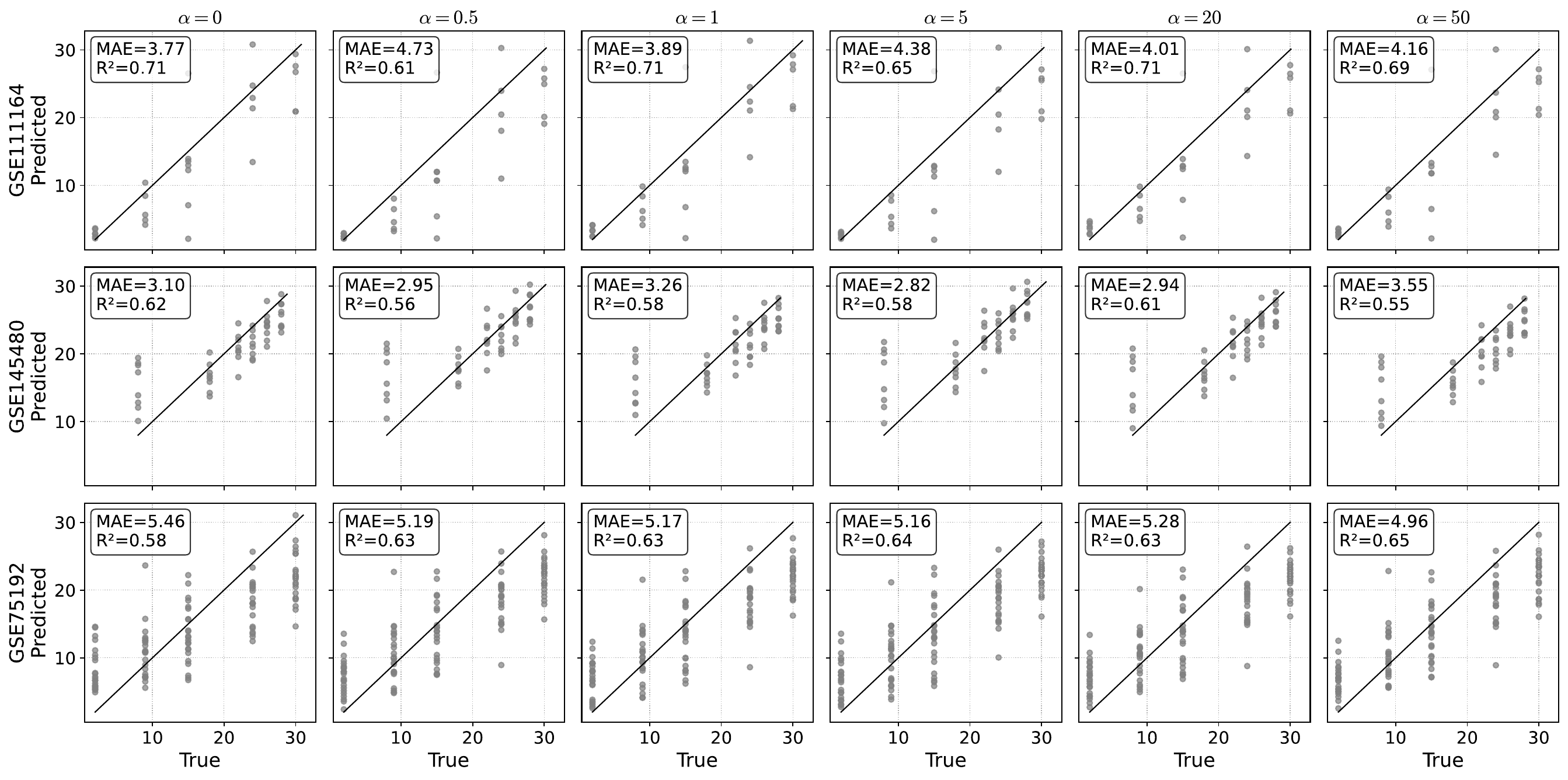}
	\caption{\textbf{Regression plots of fold-averaged predictions \emph{without the filtering layer} (Figure~\ref{fig:figure3}A) for each leave-one-dataset-out (LOSO) holdout setup.} Figure generation follows the same procedure as in Figure~\ref{fig:figure_suppl2}
}	
	\label{fig:figure_suppl3}
\end{figure}

\begin{figure}[H]
         \centering
	\includegraphics[width=\textwidth]{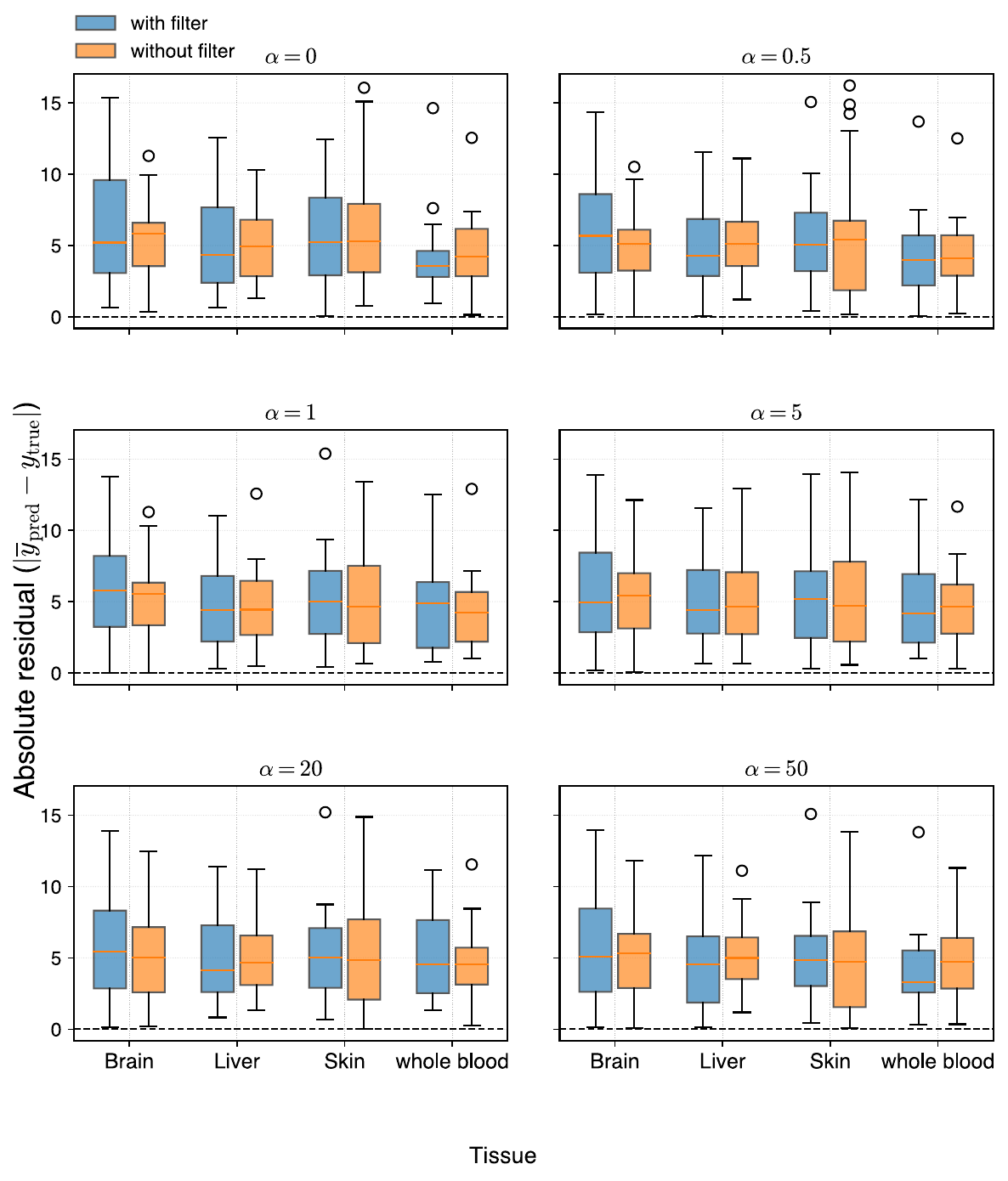}
	\caption{\textbf{Residual (absolute value) plot for by tissue fold-averaged predictions  for GSE75192 as holdout.} Predictions from all cross-validation folds were averaged for each sample before computing absolute value of residuals ($|\overline{y}_{\mathrm{pred}}-y_{\mathrm{true}}|$). Boxplots show the distribution of residuals across tissues for models trained with and without the feature filter under different values of $\alpha$
}	
	\label{fig:figure_suppl4}
\end{figure}

\bibliography{sn-bibliography}

\begin{thebibliography}{10}
\expandafter\ifx\csname url\endcsname\relax
  \def\url#1{\burl{#1}}\fi
\expandafter\ifx\csname urlprefix\endcsname\relax\def\urlprefix{URL }\fi
\providecommand{\bibinfo}[2]{#2}
\providecommand{\eprint}[2][]{\url{#2}}
\providecommand{\doi}[1]{\url{https://doi.org/#1}}
\bibcommenthead

\bibitem{pearl2009causal}
\bibinfo{author}{Pearl, J.}
\newblock \bibinfo{title}{Causal inference in statistics: An overview}.
\newblock \emph{\bibinfo{journal}{Statistics Surveys}}
  \textbf{\bibinfo{volume}{3}}, \bibinfo{pages}{96--146}
  (\bibinfo{year}{2009}).
\newblock \urlprefix\url{https://doi.org/10.1214/09-SS057}.

\bibitem{vapnik1991principles}
\bibinfo{author}{Vapnik, V.}
\newblock \bibinfo{title}{Principles of risk minimization for learning theory}.
\newblock \emph{\bibinfo{journal}{Advances in neural information processing
  systems}} \textbf{\bibinfo{volume}{4}} (\bibinfo{year}{1991}).

\bibitem{horvath2013dna}
\bibinfo{author}{Horvath, S.}
\newblock \bibinfo{title}{{DNA} methylation age of human tissues and cell
  types}.
\newblock \emph{\bibinfo{journal}{Genome biology}}
  \textbf{\bibinfo{volume}{14}}, \bibinfo{pages}{3156} (\bibinfo{year}{2013}).

\bibitem{cruz2024methylation}
\bibinfo{author}{Cruz-Gonz{\'a}lez, S.} \emph{et~al.}
\newblock \bibinfo{title}{Methylation clocks do not predict age or
  alzheimer’s disease risk across genetically admixed individuals}.
\newblock \emph{\bibinfo{journal}{bioRxiv}}  (\bibinfo{year}{2024}).

\bibitem{ferrari2022deep}
\bibinfo{author}{Ferrari, E.} \emph{et~al.}
\newblock \bibinfo{title}{A deep neural network provides an ultraprecise
  multi-tissue transcriptomic clock for the short-lived fish nothobranchius
  furzeri and identifies predicitive genes translatable to human aging}.
\newblock \emph{\bibinfo{journal}{bioRxiv}} \bibinfo{pages}{2022--11}
  (\bibinfo{year}{2022}).

\bibitem{mitchell2025mitochondria}
\bibinfo{author}{Mitchell, W.} \emph{et~al.}
\newblock \bibinfo{title}{The mitochondria-targeted peptide therapeutic
  elamipretide improves cardiac and skeletal muscle function during aging
  without detectable changes in tissue epigenetic or transcriptomic age}.
\newblock \emph{\bibinfo{journal}{Aging Cell}} \textbf{\bibinfo{volume}{24}},
  \bibinfo{pages}{e70026} (\bibinfo{year}{2025}).

\bibitem{buhlmann2020invariance}
\bibinfo{author}{B{\"u}hlmann, P.}
\newblock \bibinfo{title}{Invariance, causality and robustness}.
\newblock \emph{\bibinfo{journal}{Statistical Science}}
  \textbf{\bibinfo{volume}{35}}, \bibinfo{pages}{404--426}
  (\bibinfo{year}{2020}).

\bibitem{ying2024causality}
\bibinfo{author}{Ying, K.} \emph{et~al.}
\newblock \bibinfo{title}{Causality-enriched epigenetic age uncouples damage
  and adaptation}.
\newblock \emph{\bibinfo{journal}{Nature aging}} \textbf{\bibinfo{volume}{4}},
  \bibinfo{pages}{231--246} (\bibinfo{year}{2024}).

\bibitem{reichenbach1991direction}
\bibinfo{author}{Reichenbach, H.}
\newblock \emph{\bibinfo{title}{The direction of time}}
  Vol.~\bibinfo{volume}{65} (\bibinfo{publisher}{Univ of California Press},
  \bibinfo{year}{1991}).

\bibitem{ye2024spurious}
\bibinfo{author}{Ye, W.}, \bibinfo{author}{Zheng, G.}, \bibinfo{author}{Cao,
  X.}, \bibinfo{author}{Ma, Y.} \& \bibinfo{author}{Zhang, A.}
\newblock \bibinfo{title}{Spurious correlations in machine learning: A survey}.
\newblock \emph{\bibinfo{journal}{arXiv e-prints}} \bibinfo{pages}{arXiv--2402}
  (\bibinfo{year}{2024}).

\bibitem{vanderweele2013definition}
\bibinfo{author}{VanderWeele, T.~J.} \& \bibinfo{author}{Shpitser, I.}
\newblock \bibinfo{title}{On the definition of a confounder}.
\newblock \emph{\bibinfo{journal}{Annals of statistics}}
  \textbf{\bibinfo{volume}{41}}, \bibinfo{pages}{196} (\bibinfo{year}{2013}).

\bibitem{hernan2020causal}
\bibinfo{author}{Hern{\'a}n, M.~A.} \& \bibinfo{author}{Robins, J.~M.}
\newblock \emph{\bibinfo{title}{Causal Inference: What If}}
  (\bibinfo{publisher}{Chapman \& Hall/CRC}, \bibinfo{address}{Boca Raton},
  \bibinfo{year}{2020}).

\bibitem{luo2025bridging}
\bibinfo{author}{Luo, L.}, \bibinfo{author}{Shang, L.},
  \bibinfo{author}{Goodrich, J.~M.}, \bibinfo{author}{Peterson, K.~E.} \&
  \bibinfo{author}{Song, P.~X.}
\newblock \bibinfo{title}{Bridging the gap: Enhancing the generalizability of
  epigenetic clocks through transfer learning}.
\newblock \emph{\bibinfo{journal}{medRxiv}} \bibinfo{pages}{2025--02}
  (\bibinfo{year}{2025}).

\bibitem{watkins2023epigenetic}
\bibinfo{author}{Watkins, S.~H.} \emph{et~al.}
\newblock \bibinfo{title}{Epigenetic clocks and research implications of the
  lack of data on whom they have been developed: a review of reported and
  missing sociodemographic characteristics}.
\newblock \emph{\bibinfo{journal}{Environmental epigenetics}}
  \textbf{\bibinfo{volume}{9}}, \bibinfo{pages}{dvad005}
  (\bibinfo{year}{2023}).

\bibitem{min2024critical}
\bibinfo{author}{Min, M.}, \bibinfo{author}{Egli, C.}, \bibinfo{author}{Dulai,
  A.~S.} \& \bibinfo{author}{Sivamani, R.~K.}
\newblock \bibinfo{title}{Critical review of aging clocks and factors that may
  influence the pace of aging}.
\newblock \emph{\bibinfo{journal}{Frontiers in aging}}
  \textbf{\bibinfo{volume}{5}}, \bibinfo{pages}{1487260}
  (\bibinfo{year}{2024}).

\bibitem{koop2020epigenetic}
\bibinfo{author}{Koop, B.~E.} \emph{et~al.}
\newblock \bibinfo{title}{Epigenetic clocks may come out of
  rhythm—implications for the estimation of chronological age in forensic
  casework}.
\newblock \emph{\bibinfo{journal}{International journal of legal medicine}}
  \textbf{\bibinfo{volume}{134}}, \bibinfo{pages}{2215--2228}
  (\bibinfo{year}{2020}).

\bibitem{tomusiak2024development}
\bibinfo{author}{Tomusiak, A.} \emph{et~al.}
\newblock \bibinfo{title}{Development of an epigenetic clock resistant to
  changes in immune cell composition}.
\newblock \emph{\bibinfo{journal}{Communications Biology}}
  \textbf{\bibinfo{volume}{7}}, \bibinfo{pages}{934} (\bibinfo{year}{2024}).

\bibitem{ben2010theory}
\bibinfo{author}{Ben-David, S.} \emph{et~al.}
\newblock \bibinfo{title}{A theory of learning from different domains}.
\newblock \emph{\bibinfo{journal}{Machine learning}}
  \textbf{\bibinfo{volume}{79}}, \bibinfo{pages}{151--175}
  (\bibinfo{year}{2010}).

\bibitem{muandet2013domain}
\bibinfo{author}{Muandet, K.}, \bibinfo{author}{Balduzzi, D.} \&
  \bibinfo{author}{Sch{\"o}lkopf, B.}
\newblock \emph{\bibinfo{title}{Domain generalization via invariant feature
  representation}}, \bibinfo{pages}{10--18} (\bibinfo{organization}{PMLR},
  \bibinfo{year}{2013}).

\bibitem{gulrajani2020search}
\bibinfo{author}{Gulrajani, I.} \& \bibinfo{author}{Lopez-Paz, D.}
\newblock \bibinfo{title}{In search of lost domain generalization}.
\newblock \emph{\bibinfo{journal}{arXiv preprint arXiv:2007.01434}}
  (\bibinfo{year}{2020}).

\bibitem{sagawa2019distributionally}
\bibinfo{author}{Sagawa, S.}, \bibinfo{author}{Koh, P.~W.},
  \bibinfo{author}{Hashimoto, T.~B.} \& \bibinfo{author}{Liang, P.}
\newblock \bibinfo{title}{Distributionally robust neural networks for group
  shifts: On the importance of regularization for worst-case generalization}.
\newblock \emph{\bibinfo{journal}{arXiv preprint arXiv:1911.08731}}
  (\bibinfo{year}{2019}).

\bibitem{kuhn2025distributionally}
\bibinfo{author}{Kuhn, D.}, \bibinfo{author}{Shafiee, S.} \&
  \bibinfo{author}{Wiesemann, W.}
\newblock \bibinfo{title}{Distributionally robust optimization}.
\newblock \emph{\bibinfo{journal}{Acta Numerica}}
  \textbf{\bibinfo{volume}{34}}, \bibinfo{pages}{579--804}
  (\bibinfo{year}{2025}).

\bibitem{bai2023wasserstein}
\bibinfo{author}{Bai, X.}, \bibinfo{author}{He, G.}, \bibinfo{author}{Jiang,
  Y.} \& \bibinfo{author}{Obloj, J.}
\newblock \bibinfo{title}{Wasserstein distributional robustness of neural
  networks}.
\newblock \emph{\bibinfo{journal}{Advances in Neural Information Processing
  Systems}} \textbf{\bibinfo{volume}{36}}, \bibinfo{pages}{26322--26347}
  (\bibinfo{year}{2023}).

\bibitem{david2010impossibility}
\bibinfo{author}{David, S.~B.}, \bibinfo{author}{Lu, T.}, \bibinfo{author}{Luu,
  T.} \& \bibinfo{author}{P{\'a}l, D.}
\newblock \emph{\bibinfo{title}{Impossibility theorems for domain adaptation}},
  Vol.~\bibinfo{volume}{9}, \bibinfo{pages}{129--136}.
  \bibinfo{organization}{JMLR Workshop and Conference Proceedings}
  (\bibinfo{publisher}{PMLR}, \bibinfo{address}{Sardinia, Italy},
  \bibinfo{year}{2010}).

\bibitem{barocas2023fairness}
\bibinfo{author}{Barocas, S.}, \bibinfo{author}{Hardt, M.} \&
  \bibinfo{author}{Narayanan, A.}
\newblock \emph{\bibinfo{title}{Fairness and machine learning: Limitations and
  opportunities}}  (\bibinfo{publisher}{MIT press}, \bibinfo{year}{2023}).

\bibitem{zhao2020training}
\bibinfo{author}{Zhao, Q.}, \bibinfo{author}{Adeli, E.} \&
  \bibinfo{author}{Pohl, K.~M.}
\newblock \bibinfo{title}{Training confounder-free deep learning models for
  medical applications}.
\newblock \emph{\bibinfo{journal}{Nature communications}}
  \textbf{\bibinfo{volume}{11}}, \bibinfo{pages}{1--9} (\bibinfo{year}{2020}).

\bibitem{ganin_unsupervised_2015}
\bibinfo{author}{Ganin, Y.} \& \bibinfo{author}{Lempitsky, V.}
\newblock \bibinfo{title}{Unsupervised {Domain} {Adaptation} by
  {Backpropagation}} (\bibinfo{year}{2015}).
\newblock \urlprefix\url{http://arxiv.org/abs/1409.7495}.
\newblock \bibinfo{note}{ArXiv:1409.7495 [stat]}.

\bibitem{adeli2019representation}
\bibinfo{author}{Adeli, E.} \emph{et~al.}
\newblock \bibinfo{title}{Representation learning with statistical independence
  to mitigate bias}.
\newblock \emph{\bibinfo{journal}{arXiv preprint arXiv:1910.03676}}
  (\bibinfo{year}{2019}).
\newblock \urlprefix\url{https://arxiv.org/abs/1910.03676}.

\bibitem{roy2019mitigating}
\bibinfo{author}{Roy, P.~C.} \& \bibinfo{author}{Boddeti, V.~N.}
\newblock \emph{\bibinfo{title}{Mitigating information leakage in image
  representations: A maximum entropy approach}}, \bibinfo{pages}{2586--2594}
  (\bibinfo{publisher}{IEEE/CVF}, \bibinfo{address}{Long Beach, CA, USA},
  \bibinfo{year}{2019}).

\bibitem{elazar2018adversarial}
\bibinfo{author}{Elazar, Y.} \& \bibinfo{author}{Goldberg, Y.}
\newblock \emph{\bibinfo{title}{Adversarial removal of demographic attributes
  from text data}}, \bibinfo{pages}{11--21} (\bibinfo{year}{2018}).

\bibitem{trelin2020binary}
\bibinfo{author}{Trelin, A.} \& \bibinfo{author}{Proch{\'a}zka, A.}
\newblock \bibinfo{title}{Binary stochastic filtering: feature selection and
  beyond}.
\newblock \emph{\bibinfo{journal}{arXiv preprint arXiv:2007.03920}}
  (\bibinfo{year}{2020}).
\newblock \urlprefix\url{https://arxiv.org/abs/2007.03920}.

\bibitem{meyer2021bit}
\bibinfo{author}{Meyer, D.~H.} \& \bibinfo{author}{Schumacher, B.}
\newblock \bibinfo{title}{Bit age: A transcriptome-based aging clock near the
  theoretical limit of accuracy}.
\newblock \emph{\bibinfo{journal}{Aging cell}} \textbf{\bibinfo{volume}{20}},
  \bibinfo{pages}{e13320} (\bibinfo{year}{2021}).

\bibitem{insilico2020patent}
\bibinfo{author}{Zhavoronkov, A.} \emph{et~al.}
\newblock \bibinfo{title}{Deep aging clocks based on human transcriptomic
  data}.
\newblock \bibinfo{howpublished}{US Patent US20190034581A1}
  (\bibinfo{year}{2020}).
\newblock \urlprefix\url{https://patents.google.com/patent/US20190034581A1/en}.
\newblock \bibinfo{note}{Insilico Medicine}.

\bibitem{zakar2024profiling}
\bibinfo{author}{Zakar-Poly{\'a}k, E.}, \bibinfo{author}{Csordas, A.},
  \bibinfo{author}{P{\'a}lovics, R.} \& \bibinfo{author}{Kerepesi, C.}
\newblock \bibinfo{title}{Profiling the transcriptomic age of single-cells in
  humans}.
\newblock \emph{\bibinfo{journal}{Communications Biology}}
  \textbf{\bibinfo{volume}{7}}, \bibinfo{pages}{1397} (\bibinfo{year}{2024}).

\bibitem{Muralidharan2025.02.28.640749}
\bibinfo{author}{Muralidharan, C.} \emph{et~al.}
\newblock \bibinfo{title}{Human brain cell-type-specific aging clocks based on
  single-nuclei transcriptomics}.
\newblock \emph{\bibinfo{journal}{bioRxiv}}  (\bibinfo{year}{2025}).
\newblock
  \urlprefix\url{https://www.biorxiv.org/content/early/2025/03/02/2025.02.28.640749}.

\bibitem{costa2026multi}
\bibinfo{author}{Costa, E.~K.} \emph{et~al.}
\newblock \bibinfo{title}{Multi-tissue transcriptomic aging atlas reveals
  predictive aging biomarkers in the killifish}.
\newblock \emph{\bibinfo{journal}{Nature Aging}} \bibinfo{pages}{1--29}
  (\bibinfo{year}{2026}).

\bibitem{lopez2023hallmarks}
\bibinfo{author}{L{\'o}pez-Ot{\'\i}n, C.}, \bibinfo{author}{Blasco, M.~A.},
  \bibinfo{author}{Partridge, L.}, \bibinfo{author}{Serrano, M.} \&
  \bibinfo{author}{Kroemer, G.}
\newblock \bibinfo{title}{Hallmarks of aging: An expanding universe}.
\newblock \emph{\bibinfo{journal}{Cell}} \textbf{\bibinfo{volume}{186}},
  \bibinfo{pages}{243--278} (\bibinfo{year}{2023}).

\bibitem{naidoo2009er}
\bibinfo{author}{Naidoo, N.}
\newblock \bibinfo{title}{Er and aging—protein folding and the er stress
  response}.
\newblock \emph{\bibinfo{journal}{Ageing research reviews}}
  \textbf{\bibinfo{volume}{8}}, \bibinfo{pages}{150--159}
  (\bibinfo{year}{2009}).

\bibitem{aman2021autophagy}
\bibinfo{author}{Aman, Y.} \emph{et~al.}
\newblock \bibinfo{title}{Autophagy in healthy aging and disease}.
\newblock \emph{\bibinfo{journal}{Nature aging}} \textbf{\bibinfo{volume}{1}},
  \bibinfo{pages}{634--650} (\bibinfo{year}{2021}).

\bibitem{wu2018relevance}
\bibinfo{author}{Wu, D.} \& \bibinfo{author}{Prives, C.}
\newblock \bibinfo{title}{Relevance of the p53--mdm2 axis to aging}.
\newblock \emph{\bibinfo{journal}{Cell Death \& Differentiation}}
  \textbf{\bibinfo{volume}{25}}, \bibinfo{pages}{169--179}
  (\bibinfo{year}{2018}).

\bibitem{park2022nuclear}
\bibinfo{author}{Park, H.-S.}, \bibinfo{author}{Lee, J.}, \bibinfo{author}{Lee,
  H.-S.}, \bibinfo{author}{Ahn, S.~H.} \& \bibinfo{author}{Ryu, H.-Y.}
\newblock \bibinfo{title}{Nuclear mrna export and aging}.
\newblock \emph{\bibinfo{journal}{International Journal of Molecular Sciences}}
  \textbf{\bibinfo{volume}{23}}, \bibinfo{pages}{5451} (\bibinfo{year}{2022}).

\bibitem{angarola2021splicing}
\bibinfo{author}{Angarola, B.~L.} \& \bibinfo{author}{Anczuk{\'o}w, O.}
\newblock \bibinfo{title}{Splicing alterations in healthy aging and disease}.
\newblock \emph{\bibinfo{journal}{Wiley Interdisciplinary Reviews: RNA}}
  \textbf{\bibinfo{volume}{12}}, \bibinfo{pages}{e1643} (\bibinfo{year}{2021}).

\bibitem{harries2023dysregulated}
\bibinfo{author}{Harries, L.~W.}
\newblock \bibinfo{title}{Dysregulated rna processing and metabolism: a new
  hallmark of ageing and provocation for cellular senescence}.
\newblock \emph{\bibinfo{journal}{The FEBS Journal}}
  \textbf{\bibinfo{volume}{290}}, \bibinfo{pages}{1221--1234}
  (\bibinfo{year}{2023}).

\bibitem{saxton2017mtor}
\bibinfo{author}{Saxton, R.~A.} \& \bibinfo{author}{Sabatini, D.~M.}
\newblock \bibinfo{title}{mtor signaling in growth, metabolism, and disease}.
\newblock \emph{\bibinfo{journal}{Cell}} \textbf{\bibinfo{volume}{168}},
  \bibinfo{pages}{960--976} (\bibinfo{year}{2017}).

\bibitem{acosta2021importance}
\bibinfo{author}{Acosta-Rodr{\'\i}guez, V.~A.}, \bibinfo{author}{Rijo-Ferreira,
  F.}, \bibinfo{author}{Green, C.~B.} \& \bibinfo{author}{Takahashi, J.~S.}
\newblock \bibinfo{title}{Importance of circadian timing for aging and
  longevity}.
\newblock \emph{\bibinfo{journal}{Nature communications}}
  \textbf{\bibinfo{volume}{12}}, \bibinfo{pages}{2862} (\bibinfo{year}{2021}).

\bibitem{wolff2023defining}
\bibinfo{author}{Wolff, C.~A.} \emph{et~al.}
\newblock \bibinfo{title}{Defining the age-dependent and tissue-specific
  circadian transcriptome in male mice}.
\newblock \emph{\bibinfo{journal}{Cell reports}} \textbf{\bibinfo{volume}{42}}
  (\bibinfo{year}{2023}).

\bibitem{tyshkovskiy2024transcriptomic}
\bibinfo{author}{Tyshkovskiy, A.} \emph{et~al.}
\newblock \bibinfo{title}{Transcriptomic hallmarks of mortality reveal
  universal and specific mechanisms of aging, chronic disease, and
  rejuvenation}.
\newblock \emph{\bibinfo{journal}{Biorxiv}} \bibinfo{pages}{2024--07}
  (\bibinfo{year}{2024}).

\bibitem{cai2026learning}
\bibinfo{author}{Cai, J.} \& \bibinfo{author}{Zhu, F.}
\newblock \bibinfo{title}{Learning fair representations without labeling
  sensitive attribute via dynamic environment partitioning and invariant
  learning}.
\newblock \emph{\bibinfo{journal}{Information Processing \& Management}}
  \textbf{\bibinfo{volume}{63}}, \bibinfo{pages}{104469}
  (\bibinfo{year}{2026}).

\bibitem{akiba2019optuna}
\bibinfo{author}{Akiba, T.}, \bibinfo{author}{Sano, S.},
  \bibinfo{author}{Yanase, T.}, \bibinfo{author}{Ohta, T.} \&
  \bibinfo{author}{Koyama, M.}
\newblock \emph{\bibinfo{title}{Optuna: A next-generation hyperparameter
  optimization framework}}, \bibinfo{pages}{2623--2631} (\bibinfo{year}{2019}).

\bibitem{blitzer2007learning}
\bibinfo{author}{Blitzer, J.}, \bibinfo{author}{Crammer, K.},
  \bibinfo{author}{Kulesza, A.}, \bibinfo{author}{Pereira, F.} \&
  \bibinfo{author}{Wortman, J.}
\newblock \bibinfo{title}{Learning bounds for domain adaptation}.
\newblock \emph{\bibinfo{journal}{Advances in neural information processing
  systems}} \textbf{\bibinfo{volume}{20}} (\bibinfo{year}{2007}).

\bibitem{ganin2016domain}
\bibinfo{author}{Ganin, Y.} \emph{et~al.}
\newblock \bibinfo{title}{Domain-adversarial training of neural networks}.
\newblock \emph{\bibinfo{journal}{Journal of machine learning research}}
  \textbf{\bibinfo{volume}{17}}, \bibinfo{pages}{1--35} (\bibinfo{year}{2016}).

\bibitem{zhao2019learning}
\bibinfo{author}{Zhao, H.}, \bibinfo{author}{Des~Combes, R.~T.},
  \bibinfo{author}{Zhang, K.} \& \bibinfo{author}{Gordon, G.}
\newblock \emph{\bibinfo{title}{On learning invariant representations for
  domain adaptation}}, \bibinfo{pages}{7523--7532}
  (\bibinfo{organization}{PMLR}, \bibinfo{year}{2019}).

\bibitem{ioffe2015batch}
\bibinfo{author}{Ioffe, S.} \& \bibinfo{author}{Szegedy, C.}
\newblock \emph{\bibinfo{title}{Batch normalization: Accelerating deep network
  training by reducing internal covariate shift}}, \bibinfo{pages}{448--456}
  (\bibinfo{organization}{pmlr}, \bibinfo{year}{2015}).

\bibitem{srivastava2014dropout}
\bibinfo{author}{Srivastava, N.}, \bibinfo{author}{Hinton, G.},
  \bibinfo{author}{Krizhevsky, A.}, \bibinfo{author}{Sutskever, I.} \&
  \bibinfo{author}{Salakhutdinov, R.}
\newblock \bibinfo{title}{Dropout: a simple way to prevent neural networks from
  overfitting}.
\newblock \emph{\bibinfo{journal}{The journal of machine learning research}}
  \textbf{\bibinfo{volume}{15}}, \bibinfo{pages}{1929--1958}
  (\bibinfo{year}{2014}).

\bibitem{bishop1995training}
\bibinfo{author}{Bishop, C.~M.}
\newblock \bibinfo{title}{Training with noise is equivalent to tikhonov
  regularization}.
\newblock \emph{\bibinfo{journal}{Neural computation}}
  \textbf{\bibinfo{volume}{7}}, \bibinfo{pages}{108--116}
  (\bibinfo{year}{1995}).

\bibitem{tishby2000information}
\bibinfo{author}{Tishby, N.}, \bibinfo{author}{Pereira, F.~C.} \&
  \bibinfo{author}{Bialek, W.}
\newblock \bibinfo{title}{The information bottleneck method}.
\newblock \emph{\bibinfo{journal}{arXiv preprint physics/0004057}}
  (\bibinfo{year}{2000}).

\bibitem{krogh1991simple}
\bibinfo{author}{Krogh, A.} \& \bibinfo{author}{Hertz, J.}
\newblock \bibinfo{title}{A simple weight decay can improve generalization}.
\newblock \emph{\bibinfo{journal}{Advances in neural information processing
  systems}} \textbf{\bibinfo{volume}{4}} (\bibinfo{year}{1991}).

\end{thebibliography}

\end{document}